%% file: main.tex
\documentclass[nohyperref]{article}

\usepackage{microtype}
\usepackage{graphicx}
\usepackage{subfigure}
\usepackage{booktabs} % for professional tables

\usepackage[accepted]{icml2023}

\usepackage{amsmath}
\usepackage{amssymb}
\usepackage{mathtools}
\usepackage{amsthm}
\usepackage{bm}
\usepackage{overpic}
\usepackage{makecell}
\usepackage{float}

\definecolor{citeblue}{RGB}{48,111,186}
\definecolor{c1}{HTML}{fe793d}
\definecolor{c2}{HTML}{b71a3b}
\definecolor{c3}{HTML}{7e0f12}
\usepackage[pagebackref=false,breaklinks=true,colorlinks=true,citecolor=citeblue,bookmarks=false]{hyperref}
\usepackage[capitalize,noabbrev]{cleveref}
\usepackage{booktabs,multirow,adjustbox,diagbox,threeparttable,arydshln}

\crefname{section}{Sec.}{Secs.}
\Crefname{section}{Section}{Sections}
\crefname{table}{Tab.}{Tabs.}
\Crefname{table}{Table}{Tables}
\crefname{figure}{Fig.}{Figs.}
\Crefname{figure}{Figure}{Figures}
\crefname{equation}{Eq.}{Eqs.}
\Crefname{equation}{Equation}{Equations}
\Crefname{conjecture}{Conjecture}{Conjectures}
\usepackage{apptools}
\crefname{subappendix}{\IfAppendix{Sec.}{appendix}}{\IfAppendix{Secs.}{appendices}s}

\protected\def\ignorethis#1\endignorethis{}
\let\endignorethis\relax

\def\TOCstart{\addtocontents{toc}{\endignorethis}}

\theoremstyle{plain}
\newtheorem{theorem}{Theorem}[section]

\theoremstyle{definition}

\theoremstyle{remark}

\usepackage[textsize=tiny]{todonotes}
\usepackage{capt-of}
\usepackage{xspace}

\newcommand{\methodbar}{Cones\xspace}
\newcommand{\etcno}{\textit{etc.}}
\newcommand{\ieno}{\textit{i.e.}}
\newcommand{\vepsilon}{\bm{\epsilon}}
\newcommand{\vtheta}{\bm{\theta}}
\newcommand{\diff}{\hat{\vx}_{\vtheta}}
\newcommand{\para}{\bm{\Theta}}
\newcommand{\con}{\gL_{\rm con}}

\input{math_cmd.tex}

\icmltitlerunning{Cones: Concept Neurons in Diffusion Models for Customized Generation}

\begin{document}

\twocolumn[{
\icmltitle{Cones: Concept Neurons in Diffusion Models for Customized Generation}



\icmlsetsymbol{equal}{*}
\icmlsetsymbol{intern}{$\ddag$}
\icmlsetsymbol{corresponding}{$\dagger$}

\begin{icmlauthorlist}
\icmlauthor{Zhiheng Liu}{yyy,equal,intern}
\icmlauthor{Ruili Feng}{yyy,equal,intern}
\icmlauthor{Kai Zhu}{yyy}
\icmlauthor{Yifei Zhang}{comp,intern}
\icmlauthor{Kecheng Zheng}{sch}
\icmlauthor{Yu Liu}{ooo}
\icmlauthor{Deli Zhao}{ooo}
\icmlauthor{Jingren Zhou}{ooo}
\icmlauthor{Yang Cao}{yyy,corresponding}
\end{icmlauthorlist}

\icmlaffiliation{yyy}{University of Science and Technology of China}
\icmlaffiliation{comp}{Shanghai Jiao Tong University}
\icmlaffiliation{sch}{Ant Group}
\icmlaffiliation{ooo}{Alibaba Group}

\icmlcorrespondingauthor{Zhiheng Liu}{lzh990528@mail.ustc.edu.cn}
\icmlcorrespondingauthor{Ruili Feng}{ruilifengustc@gmail.com}
\icmlcorrespondingauthor{Kai Zhu}{zkzy@mail.ustc.edu.cn}
\icmlcorrespondingauthor{Yifei Zhang}{qidouxiong619@sjtu.edu.cn}
\icmlcorrespondingauthor{Kecheng Zheng}{zkechengzk@gmail.com}
\icmlcorrespondingauthor{Yu Liu}{ly103369@alibaba-inc.com}
\icmlcorrespondingauthor{Deli Zhao}{zhaodeli@gmail.com}
\icmlcorrespondingauthor{Jingren Zhou}{jingren.zhou@alibaba-inc.com}
\icmlcorrespondingauthor{Yang Cao}{forrest@ustc.edu.cn}
% \renewcommand\twocolumn[1][]{#1}%
% \vspace{-40pt}
% \begin{center}
%   \centering
%   \includegraphics[width=1\textwidth]{fig/fig1.pdf}
%   \vspace{-20pt}
%   \captionof{figure}{fig1}
% \end{center}

\renewcommand\twocolumn[1][]{#1}

\thispagestyle{empty}
\begin{center}
  \centering
  \begin{overpic}[width=1\linewidth]{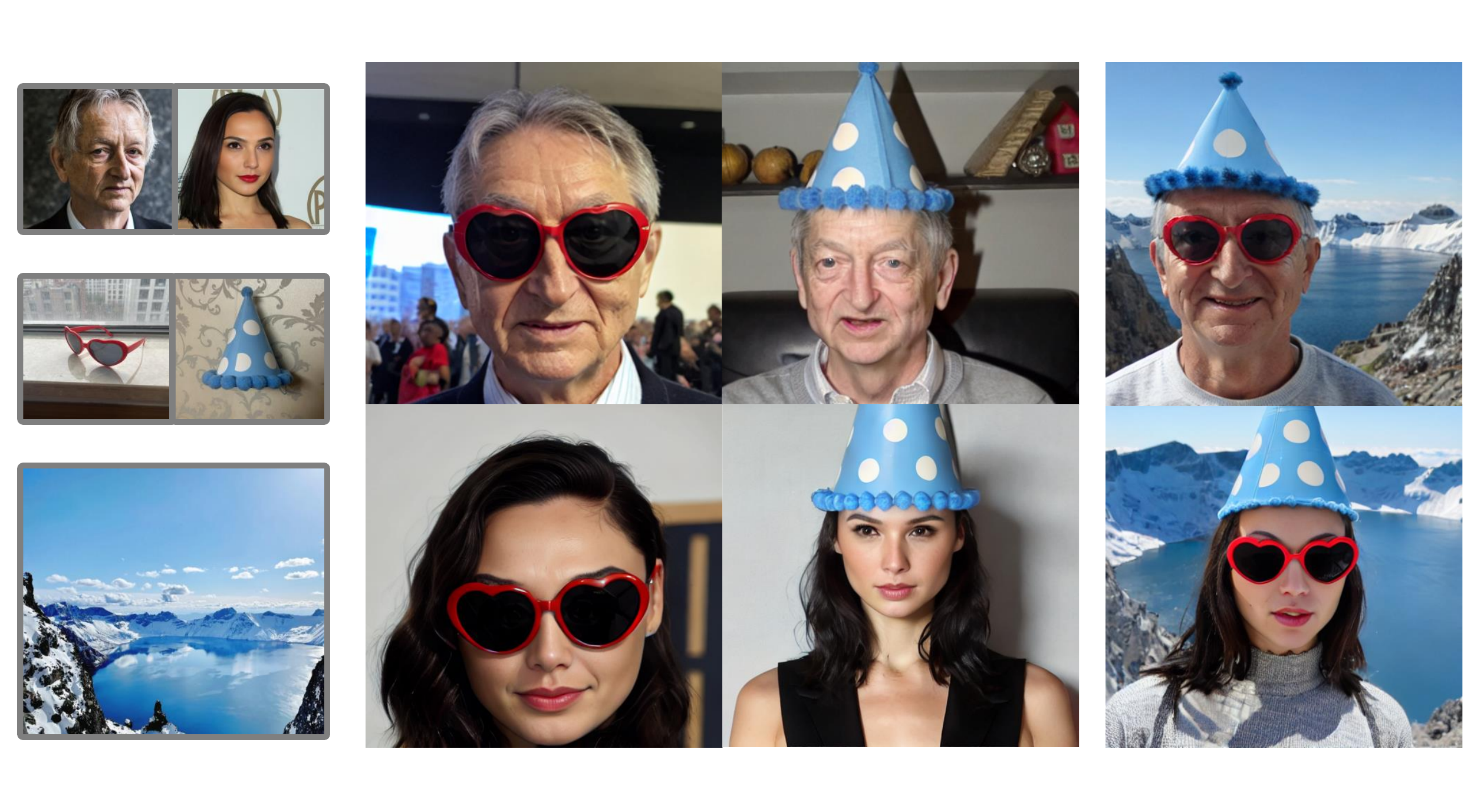}\vspace{-0.4cm}
  \put(5,50){Subject 1 (Person)}
  % \put(53,52){Generated Images}
   \put(5,37){Subject 2 (Object)}
   \put(3,24){Subject 3 (Background)}
   \put(3,2){(a) Reference images}
   \put(30,2){(b) Images generated by concatenating subject-specific concept neurons}
   \put(42,51.5){Person + Object}
   \put(73.5,51.5){\makecell{Person + Objects + Background}}
%   \put(25.7,17){\rotatebox{90}{Concept 2 (Person)}}
%   \put(5,59){\makecell{Text Prompt: \\
% \textcolor{c3}{A V1* man/woman,} \\
% \textcolor{c2}{with V2* object} \\
% \textcolor{c2}{on the face,} \\
% \textcolor{c1}{with V3* lake} \\
% \textcolor{c1}{in the background.}}}
%   \put(42,47){Concept 1 (Object)}
%   \put(77,47){Concept 3 (Background)}
  \end{overpic}
  \captionof{figure}{We explore the subject-specific concept neurons in a pre-trained text-to-image diffusion model. Concatenating
multiple clusters of concept neurons representing different persons, objects, and backgrounds can flexibly generate all related concepts in a single image.}\label{fig:teaser}
\end{center}
% \begin{center}
%   \centering
%   \begin{overpic}[width=1\linewidth]{fig/fig1.pdf}\vspace{-0.4cm}
%   \put(25.7,17){\rotatebox{90}{Concept 2 (Person)}}
%   \put(5,59){\makecell{Text Prompt: \\
% \textcolor{c3}{A V1* man/woman,} \\
% \textcolor{c2}{with V2* object} \\
% \textcolor{c2}{on the face,} \\
% \textcolor{c1}{with V3* lake} \\
% \textcolor{c1}{in the background.}}}
%   \put(42,47){Concept 1 (Object)}
%   \put(77,47){Concept 3 (Background)}
%   \end{overpic}
%   \captionof{figure}{We explore the subject-specific concept neurons in a pre-trained text-to-image diffusion model. Concatenating
% multiple clusters of concept neurons representing different persons, objects, and backgrounds can flexibly generate all related concepts in a single image.}\label{fig:teaser}
% \end{center}

\icmlkeywords{Machine Learning, ICML}
\vskip 0.3in
}]

% \vspace{-0.7cm}
% \begin{figure*}[t]
% \centering
% \includegraphics[width=\textwidth]{fig/fig1.pdf}
% \caption{We explore the subject-specific concept neurons in a pre-trained text-to-image diffusion model. Concatenating
% multiple clusters of concept neurons representing different persons, objects and background can flexibly generate all related concepts in a single image.} 
% \label{fig:teaser} 
% \end{figure*}

% \twocolumn[{
% \renewcommand\twocolumn[1][]{#1}

% \thispagestyle{empty}
% \begin{center}
%   \centering
%   \includegraphics[width=1\linewidth]{fig/fig1.pdf}\vspace{-0.4cm}
%   \captionof{figure}{Given a set of commercial models with underwear and different clothing images, the proposed method can generate realistic clothing model results with clear pattern reconstruction.}\label{fig:teaser}
% \end{center}
% }]



%\printAffiliationsAndNotice{am}  % leave blank if no need to mention equal contribution
\printAffiliationsAndNotice{\icmlEqualContribution. $^\ddag$Work performed during internship at Alibaba DAMO Academy. $^\dagger$Corresponding Author} % otherwise use the standard text.

% \vspace{-0.7cm}
% \begin{figure*}[H]
% \centering
% \includegraphics[trim={0 0 0 0},clip,width=\textwidth]{fig/fig1.pdf}
% \caption{We explore the subject-specific concept neurons in a pre-trained text-to-image diffusion model. Concatenating
% multiple clusters of concept neurons representing different persons, objects and background can flexibly generate all related concepts in a single image.} 
% \label{fig:teaser} 
% \end{figure*}

%\input{section/teaser.tex}
\input{section/0.abs.tex}
\input{section/1.intro.tex}
\input{section/3.method.tex}
\input{section/4.exps.tex}

\input{section/5.conclusion.tex}
\input{section/6.ref.tex}
\clearpage
\TOCstart
\input{section/7.appendix.tex}

\end{document}

%% file: math_cmd.tex
\usepackage{amsmath,amsfonts,bm}

\newcommand{\myparagraph}[1]{\vspace{1pt} \noindent \textbf{#1} \ }
\def\eqref#1{equation~\ref{#1}}
\def\1{\bm{1}}

\def\vtheta{{\bm{\theta}}}

\def\vc{{\bm{c}}}

\def\vx{{\bm{x}}}

\def\mM{{\bm{M}}}

\def\mV{{\bm{V}}}

\DeclareMathAlphabet{\mathsfit}{\encodingdefault}{\sfdefault}{m}{sl}
\SetMathAlphabet{\mathsfit}{bold}{\encodingdefault}{\sfdefault}{bx}{n}

\def\gH{{\mathcal{H}}}
\def\gI{{\mathcal{I}}}

\def\gL{{\mathcal{L}}}

\def\gN{{\mathcal{N}}}
\def\gO{{\mathcal{O}}}

\def\gX{{\mathcal{X}}}

\def\sR{{\mathbb{R}}}

\newcommand{\pdata}{p_{\rm{data}}}
\newcommand{\E}{\mathbb{E}}

%% file: section/0.abs.tex
\begin{abstract}
Human brains respond to semantic features of presented
stimuli with different neurons.
It is then curious whether modern deep neural networks admit a similar behavior pattern.
Specifically, this paper finds a small cluster of neurons in a diffusion model corresponding to a particular subject.
We call those neurons the concept neurons.
They can be identified by statistics of network gradients to a stimulation connected with the given subject.
The concept neurons demonstrate magnetic properties in interpreting and manipulating generation results. 
Shutting them can directly yield the related subject contextualized in different scenes.
Concatenating multiple clusters of concept neurons can vividly generate all related concepts in a single image.
A few steps of further fine-tuning can enhance the multi-concept capability, which may be the first to manage to generate up to four different subjects in a single image.
For large-scale applications, the concept neurons are environmentally friendly as we only need to store a sparse cluster of int index instead of dense float32 values of the parameters, which reduces storage consumption by 90\% compared with previous subject-driven generation methods.
Extensive qualitative and quantitative studies on diverse scenarios show the superiority of our method in interpreting and manipulating diffusion models.
Our code is available at \href{https://github.com/Johanan528/Cones} {https://github.com/Johanan528/Cones}.
\end{abstract}

%% file: section/1.intro.tex
\section{Introduction}\label{sec:intro}
The sophisticated structure of human brains allows miraculous cognitive and imaginative capabilities. Research has found that concept neurons in the human medial temporal lobe respond to semantic features of different presented stimuli separately \cite{bausch2021concept, thiebaut2022emergent}. Those neurons encode temporal as well as abstract relations among elements of experience across spatiotemporal gaps and are thought to be the key to high-level intelligence~\cite{bausch2021concept}.

It is then curious to know, as one of the most successful artificial intelligence systems, do modern deep neural networks~\cite{lecun2015deep} admit a similar structure of concept neurons. Specifically, to mimic the imaginative ability of the human brain, do generative diffusion models \cite{ho2020denoising, dhariwal2021diffusion} encode different subjects separately with their neurons? This paper is about to answer this question from the perspective of subject-driven generation~\cite{cosdiff,dreambooth}. We propose to find a small cluster of neurons, which are parameters in the attention layer of a pretrained text-to-image diffusion model \cite{rombach2022high}, such that changing values of those neurons can generate a corresponding subject in different contents, based on the semantics in the input text prompt. We attribute these neurons as the concept neurons connected to the corresponding subject in the diffusion models. Finding them can advance our understanding of the underlying mechanism of deep diffusion networks and provide an original methodology for subject-driven generation.

This paper proposes a novel gradient-based method to analyze and identify the concept neurons, termed as Cones\footnote{\textbf{Cones} (\textbf{Co}ncept \textbf{ne}uron\textbf{s}) is inspired by the name of photoreceptor cells in the retina of the eye known as ``cone cells''~\cite{enwiki:1141645734}.}. We motivate them as the parameters that scale down whose absolute value can better construct the given subject while preserving prior information. This motivation can then induce a gradient-based criterion for whether a parameter is a concept neuron. Using this criterion, we can find all the concept neurons after a few gradient computations.

We then study the interpretability of those concept neurons from several perspectives. We first investigate the robustness of concept neurons to changes in their values. We optimize a concept-implanting loss~\cite{dreambooth} on the concept neurons using float32, float16, quaternary, and binary (shutting those concept neurons directly without training) digital accuracy correspondingly. The results show similar performance among all the settings, demonstrating the strong robustness of concept neurons in controlling the target subject. While binary digital accuracy requires no further training and minimum storage space, we use it as our default subject-driven generation method. This method further admits fascinating additivity---concatenating concept neurons of multiple subjects directly can generate them all in the results, which may be the first to discover such a simple yet effective affine semantic structure in the parameter space of diffusion models. Further fine-tuning based on the concatenating can promote the multi-concept generation capability to a new milestone: we are the first to manage to generate four different diverse subjects in one image in the domain of subject-driven generation. Finally, thanks to their sparsity and robustness, the concept neurons can be efficiently used in large-scale applications. Storing the information to construct a given subject costs around only 10\% of memory compared with previous subject-driven methods~\cite{dreambooth,cosdiff}, which is extremely economical and environment-friendly for commercial usage in mobile devices. Extensive studies on diverse categories, ranging from human portraits, scenes, decorations, \etcno, demonstrate the superiority of over method in interpretability and multi-concept generation capability.

%% file: section/3.method.tex
\section{Preliminaries and Background}\label{sec:pre}
\myparagraph{Diffusion Models.} Diffusion models~\cite{ho2020denoising, dhariwal2021diffusion, rombach2022high,song2020score} are parametric neural networks that learn image distributions by gradual denoising. To further explore the extensibility of diffusion models, many works have been devoted to diffusion-based conditional generation, which can be broadly classified into two categories. The first one is the approach known as classifier-guidance \cite{liu2023more}, which utilizes a classifier to promote the sampling process of the pre-trained unconditional model. Despite the low cost, the generation effect is less competitive. The second one is known as the classifier-free approach \cite{ho2022classifier}, which directly collects a large amount of data pairs for joint optimization under the guarantee of conditional probabilistic derivation. This approach can yield stunningly detailed results but requires a huge amount of data and computation resources. Owing to advances in language \cite{radford2021learning} and cross-modal foundation models \cite{radford2021learning}, much text-to-image work~\cite{saharia2022photorealistic, ramesh2022hierarchical, nichol2021glide} with classifier-free techniques is beginning to emerge, facilitating explicit control on the corresponding semantics and style. However, the expressiveness of text is still limited, and more work wants to utilize additional, conditional information (\textit{e.g.}, reference image, grounding \cite{li2023gligen} and sketch \cite{voynov2022sketch}) to guide the global control further. 

\myparagraph{Text-to-Image Diffusion Model.} A text-to-image diffusion model~\cite{yu2022scaling,saharia2022photorealistic,rombach2022high} $\diff$ will guide this denoising procedure with a text prompt describing the image content. Typically, it is trained by denoising a noised image $\vx_t=\alpha_t\vx+\sigma_t\vepsilon$ as
\begin{equation}
\E_{(\vx,\vc)\sim\pdata,t,\vepsilon}[\omega_t\Vert\diff(\vx_t,t,\vc)-\vx\Vert_2^2].
\end{equation}
Here $(\vx,\vc)$ are (image, text prompts) pairs sampled from data; $\vepsilon$ are standard Gaussian noise added to the noised image; $\alpha_t,\omega_t$, and $\sigma_t$ are hyper-parameter scalars to control the noise schedule evolved by time variable $t$ from $0,1,\cdots,T$. After training, the model $\diff$ can generate various images described by the text prompts by denoising standard Gaussian noises. Throughout this work, we use Stable Diffusion V1.4~\cite{rombach2022high} as the default text-to-image diffusion model due to its state-of-the-art performance and easy availability. However, Cones can also be simply applied to most text-to-image diffusion models like Imagen~\cite{saharia2022photorealistic} and DALLE-2~\cite{ramesh2022hierarchical}.

\myparagraph{Customized Generation.}
The purpose of customized generation, as first proposed in DreamBooth~\cite{dreambooth}, is to implant a given subject into the diffusion model and bind it with a unique text identifier to indicate its presence; so that the model can generate various renditions of the subject vividly guided by text prompts~\cite{lu2020countering,lee2019countering}. To capture the subject $\gX$, we need a few (usually 3 to 5) images of this subject $\{\gX^i\}_{i=1}^s$ causally taken from different point-views and conditions. As in previous work~\cite{dreambooth}, the subject $\gX$ can be implanted to the diffusion model $\diff$ by minimizing a concept-preserving loss $\gL_{\rm con}$ together with a prior-preserving loss $\gL_{\rm pr}$ in the parameter space $\vtheta\in\para$. Let $\gX^i_t=\alpha_t\gX^i+\sigma_t\vepsilon,\vepsilon\sim\gN(\gO,\gI)$, and $\vc^{\gX}$ be the text prompt `A $\mV^*$ [category name]' for this subject, then the subject-preserving loss is
\begin{equation}
    \gL_{\rm sub}=\E_{\gX^i,\vepsilon,t}[\omega_t\Vert\diff(\gX^i_t,t,\vc^{\gX})-\gX^i\Vert_2^2].
\end{equation}
It explicitly binds the subject with the text identifier $\mV^*$. To avoid over-fitting and language-drift \cite{dreambooth}, we further need the prior-preserving loss
\begin{gather}
    \gL_{\rm pr}=\E_{(\vx^{\rm pr},\vc^{\rm pr}),\vepsilon,t}[\omega_t\Vert\diff(\vx^{\rm pr}_t,t,\vc^{\rm pr})-\vx^{\rm pr}\Vert_2^2],
\end{gather}
where $(\vx^{\rm pr},\vc^{\rm pr})$ are image text prompt pairs with different subjects but the same category as the subject to implant. The full objective function is a combination of them as
\begin{equation}\label{eq:combine_loss}
    \gL_{\rm con}=\gL_{\rm sub}+\lambda\gL_{\rm pr}.
\end{equation}

\section{Method}\label{sec:method}
Our purpose here is to locate the corresponding neurons that control the generation of the given subject and use those neurons to guide customized generation~\cite{dreambooth,cosdiff}. For a diffusion model $\diff,\vtheta=(\vtheta_1,\cdots,\vtheta_n)^T\in\para\subset\sR^N$, where $N$ is the parameter volume, we want to find a small collection of neurons $\vtheta_{\gH}=(\vtheta_{h_1},\cdots,\vtheta_{h_n})^T, 1\leq h_1<h_2<\cdots< h_n\leq N,n\ll N,$ so that changes in them alone is enough to produce renditions of the given subject in different contexts, based on the text prompts. 

This task significantly differs from previous customized generation work and is much more challenging. Here we not only pursue the generation quality of the subject but are also eager for the underline mechanism of how the diffusion model memorizes subjects in its parameter space. So our primary focus is on the interpretability of network neurons and how they influence the generation.

\myparagraph{Advantages.}Such methodology will allow significant practical advantages. As we will show in~\cref{sec:locate}, simple statistics of network gradients can efficiently identify those concept neurons. Once locating them, we can \emph{directly add} concept neurons of multiple subjects to generate them all together in the results, demonstrating the powerful interpretability of Cones as is discussed in~\cref{sec:inte}. A couple of fine-tuning steps based on the above addition results can further enhance the generation in visual quality and multi-subject capability. To the best of our knowledge, this is the first method to manage to generate four different diverse subjects in one image. We will discuss this in detail in~\cref{sec:coll}. Another superiority of Cones is its storage efficiency. Thanks to the sparsity and robust binary representation of concept neurons, storing concept neurons consumes only around 10\% memory of Custom Diffusion~\cite{cosdiff} and 0.05\% memory of Dreembooth~\cite{dreambooth}. This storage-friendly property enables large-scale commercial applications of customized generation.

\subsection{Concept Neurons for a Given Subject}\label{sec:locate}
In this section, we analyze how neurons in a text-to-image diffusion model react to different subjects and locate the concept neurons corresponding to a given subject. While previous research shows that the K-V attention layer~\cite{cosdiff} dominates the subject generation process, we follow its setting and limit the search for concept neurons in the parameters of the K-V attention layers. In what follows, we always assume that $\Theta$ is the parameter family for K-V attention layers.

Cones is inspired by Functional magnetic resonance imaging (fMRI)~\cite{huettel2004functional} in medicine. It measures the small changes in blood flow that occur with brain activity. Research believes that different concepts, like visual objects or elements of experience, will induce blood-oxygen-level-dependent contrast~\cite{logothetis2001neurophysiological,kwong1992dynamic,sharoh2019laminar} in brain neurons of the human medial temporal lobe separately. Those brain neurons that prefer a specific concept are called brain concept neurons~\cite{bausch2021concept} for this concept, and they dominate the brain response to this concept. We thus wonder whether there is also a preference for concept in the neurons of a pretrained diffusion model and whether they are responsible for generating this concept.

Specifically, we want to find a couple of neurons that, scaling down their absolute value, can reconstruct the subject while maintaining prior information of the model, thus being able to generate the given subject in diverse contexts. This is equivalent to decreasing the value of the concept-implanting loss in~\cref{eq:combine_loss}. We do not consider the effect of scaling up for numerical reasons (values of scaling up can be up to infinity while scaling down is bounded by the initial values of neurons). Let $\theta=\vtheta_h$ denote the $h$-element of the whole parameter vector $\vtheta$ for simplicity. Scaling it by factor $\alpha$ will produce concept-implanting loss $\con(\alpha\theta)$. Let $\rho=(1-\alpha)(\theta\frac{\partial\con}{\partial\theta})^{-1}$. We can rewrite it as
\begin{equation}\label{eq:scale}
    \con(\alpha\theta)=\con(\theta(1-\rho\theta\frac{\partial\con}{\partial\theta})).
\end{equation}
Through Taylor expansions~\cite{rudin1976principles}, we know that 
\begin{equation}\label{eq:taylor}
    \begin{aligned}
    \con(\alpha\theta)\approx\con(\theta)-\rho\theta^2\frac{\partial\con}{\partial\theta}^2<\con(\theta)
    \end{aligned}
\end{equation}
as long as $0<\rho\ll1$. To make~\cref{eq:scale} is a scaling down, we need (when $0<\rho\ll1$)
\begin{equation}\label{eq:equiv}
    0<\alpha=1-\rho\theta\frac{\partial\con}{\partial\theta}<1\Leftrightarrow \theta\frac{\partial\con}{\partial\theta}>0.
\end{equation}
In conclusion, $\theta\frac{\con}{\partial\theta}>0$ will indicate whether scaling down the $h$-th parameter will decrease the concept-implanting loss and thus identify whether $\theta_h$ is a concept neuron for the given subject $\gX$. Rigorously, we can have the following theorem.
\begin{theorem}[Identification of Concept Neurons]\label{th}
For a given parameter $\theta\in\vtheta$, slightly scaling down it can decrease the concept-implanting loss, which is equivalent to
\begin{equation}
    \theta\frac{\partial\con}{\partial\theta}>0,
\end{equation}
and the decreasing value is proportional to $(\theta\frac{\partial\con}{\partial\theta})^2$. Thus $\theta$ is a concept neuron if and only if $\theta\frac{\partial\con}{\partial\theta}>0$.
\end{theorem}
Following this theorem, a naive method to detect whether $\theta\in\vtheta$ is a concept neuron can be that we sample $K$ different values $\theta^1,\cdots,\theta^K$ ranging from zero to $\theta$, and if 
\begin{equation}
    \theta^1\frac{\partial\con}{\partial\theta}(\theta^1)+\cdots\theta^K\frac{\partial\con}{\partial\theta}(\theta^K)>\tau>0,
\end{equation}
where $\frac{\partial\con}{\partial\theta}(\theta^k)$ is the gradient at point $\theta=\theta^i,k\in[K]$ and $\tau$ is a constant hyper-parameter, then $\theta$ is a concept neuron. 

We deduce a self-adaptive sampling method for the choices of $\theta^1,\cdots,\theta^K$. We set $\theta^1=\theta$, and 
\begin{equation}\label{eq:sampling}
    \theta^{k+1}=\theta^k(1-\rho\theta^k\frac{\partial\con}{\partial\theta}(\theta^k)),k=1,\cdots,K-1.
\end{equation}
It will sample more densely in the neighborhood where $\vert\theta\frac{\partial\con}{\partial\theta}\vert$ is small, thus ambiguous to indicate the valuence, while sparsely when the valuence is obvious. Parameters with more ambiguous regions will tend to be excluded from concept neurons. Thus the identification of concept neurons will be more cautious and robust.

For all the parameters $\vtheta$ of the diffusion model $\diff$, we can use~\cref{algorithm:CN} to compute a concept neuron mask parallelistically to indicate whether each neuron is or not a concept neuron. The main computation~\cref{eq:main} can be further accelerated using the Newton-Leibniz law of the calculus~\cite{rudin1976principles}. We use this accelerated version in practice. See appendix for detail.

\begin{algorithm}[ht]
	\caption{Computing Concept Neuron Mask}
	\label{algorithm:CN}
	\begin{algorithmic}
		\STATE {\bfseries Input:} Concept-implanting loss function $\con$, parameter $\vtheta\in\sR^n$, maximum sampling number $K$, hyper-parameter $0<\rho\ll1$ and $\tau>0$.
		\STATE {\bfseries Set:} $k=1$ and $\vtheta^k=\vtheta$.
		\REPEAT
		        \STATE compute 
          \begin{equation}\label{eq:main}
              \vtheta^{k+1}=\vtheta^k\odot(\bm{1}-\rho\vtheta^k\odot\nabla_{\vtheta}\con(\vtheta^k));
          \end{equation}
		\STATE update $k=k+1$;
		\UNTIL{ $k=K-1$.}
    \STATE {\bfseries Compute: } $\mM_p=\vtheta^1\odot\nabla_{\vtheta}\con(\vtheta^1)+\cdots+\vtheta^K\odot\nabla_{\vtheta}\con(\vtheta^K)$;
    \STATE{\bfseries Set:} $\mM=\bm{1}-(\mM_p>\tau)$.
	\STATE {\bfseries Output:} Binary concept neuron mask $\mM$ to indicate whether each neuron is a concept neuron, 1 for not and 0 for is.
	\end{algorithmic}
\end{algorithm}

\subsection{Interpretability of Concept Neurons}
In this section, we explore various aspects of the interpretability of concept neurons. Based on our findings, we propose a new customized generation method, \methodbar, which implants the subject into a diffusion model by shutting the corresponding concept neurons.

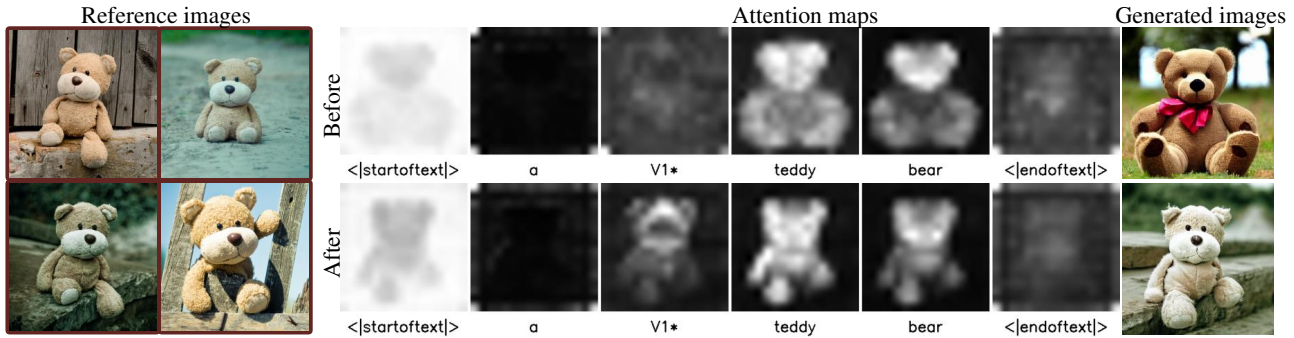
\begin{figure*}[t]
  \centering
    \begin{overpic}[width=0.99\linewidth]{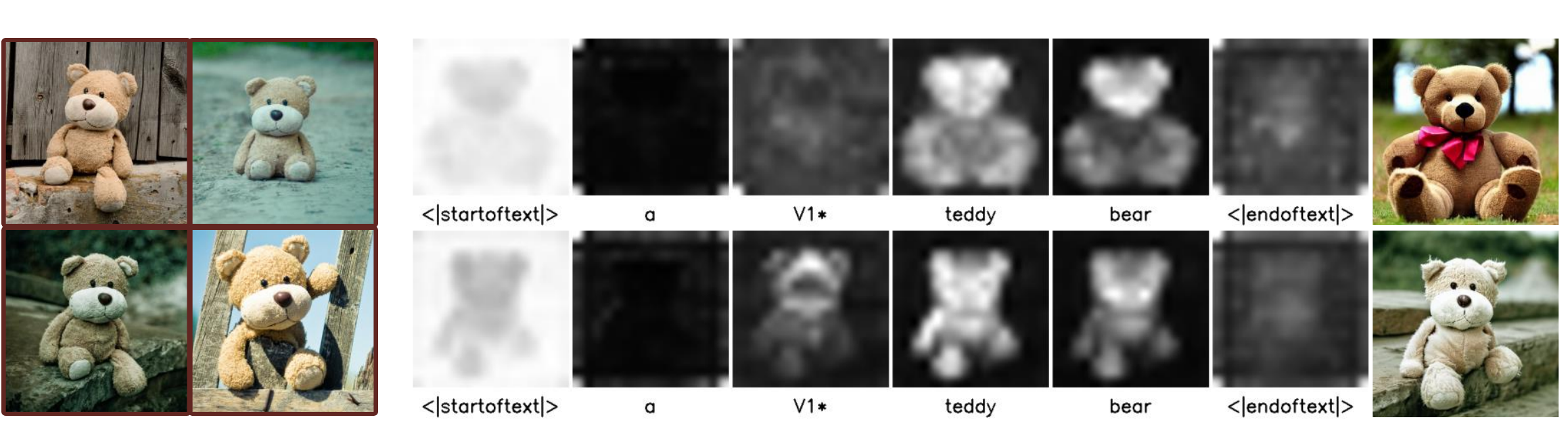}
    \put(25,16){\rotatebox{90}{\small{Before}}}
    \put(25,5){\rotatebox{90}{\small{After}}}
  \put(6,25){\small{Reference images}}
  \put(57,25){\small{Attention maps}}
  \put(87,25){\small{Generated images}}
  \end{overpic}
 
  \caption{Attention maps before and after shutting subject-specific concept neurons. Shutting concept neurons draws the outline of given subject at the attention map of the text identifier $V1*$.}

\label{fig:attention}
\end{figure*}
  
\myparagraph{Concept Neurons Indeed Responsible for Generation of the Corresponding Subject.} In~\cref{fig:attention}, which shows the attention map of the diffusion model $\diff$ before and after shutting the concept neurons corresponding to the given subject. We visualize the attention of each word in the text prompt. Shutting the concept neurons immediately draw the outline of the given subject in the attention map corresponding to the text identifier and subsequently generate the subject in the final output. This shows the strong connections between concept neurons and the given subject in the network representations.

\begin{figure}[t]
  \centering
    \begin{overpic}[width=0.99\linewidth]{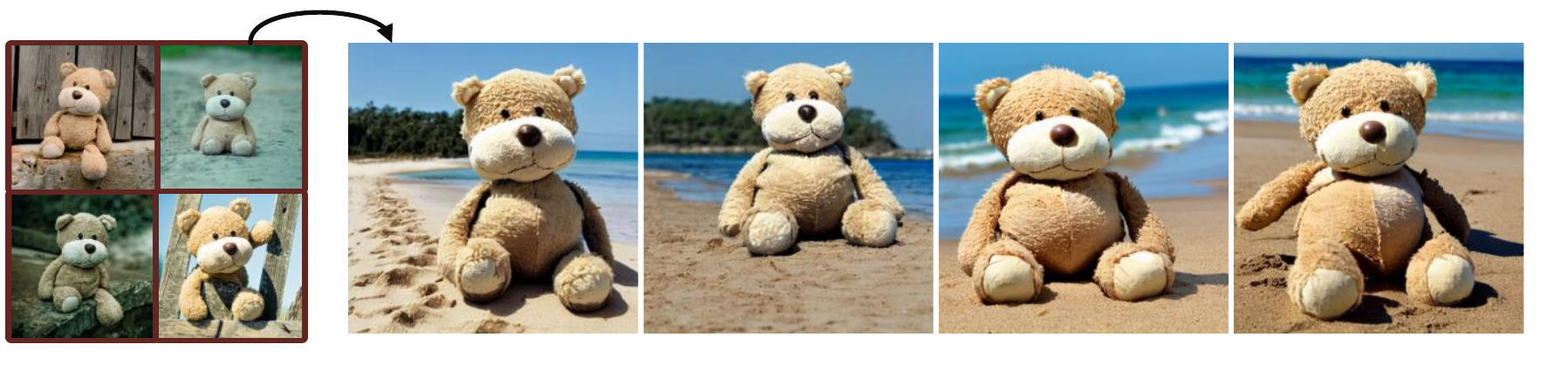}
  \put(3,-1.5){\small{Reference}}
  \put(27,-1.5){\small{Binary}}
  \put(42,-1.5){\small{Quaternary}}
  \put(64,-1.5){\small{Float16}}
  \put(83,-1.5){\small{Float32}}
  \put(8,23){\small{On a sunny day, the $V_{1}*$ teddybear is sitting on the beach.}}
  % \put(68,33){\small{\makecell{A new1 cat playing with \\ a new2 wooden pot.}}}
  \end{overpic}

  \caption{Results of optimizing the concept-implanting loss on concept neurons with float32, float16, quaternary, and binary digital accuracy. Binary digital accuracy corresponds to shutting the concept neuron without any further tuning. We can observe close performance for all cases. This demonstrates the strong robustness of controlling the subject generation of concept neurons.}
  \label{fig:moti_float}
  
\end{figure}

\myparagraph{Float32, Float16, Quaternary, and Binary Changes to the Concept Neurons Report Equal Effects.} By our motivation, changes to concept neurons should decrease the concept-implanting loss and thus generate the given subject. To demonstrate the strong interpretability and robustness of concept neurons, we study the effects of changes in different digital accuracy. We optimize the concept-implanting loss on the concept neurons and freeze the remains. We set the digital accuracy of the optimization to float32, float16, quaternary, and binary, in turn. We find close performance in all those cases, as is shown in~\cref{fig:moti_float}. This demonstrates the strong robustness of concept neurons controlling the generation of the target subject. 

Due to the strong performance and interpretability of binary digital accuracy, we name it \textbf{\emph{\methodbar}} for customized generation and use it as our default method to implant the subject into diffusion models with concept neurons. Under this setting, we will first compute the concept neuron mask for a given subject and directly multiply it to the network parameter $\vtheta$. The modified network $\hat{\vx}_{\mM\odot\vtheta}$ can then generate the target subject. This omits any further optimization on the concept neurons and is thus efficient and robust.  

\begin{figure}[t]
  \centering
    \begin{overpic}[width=0.99\linewidth]{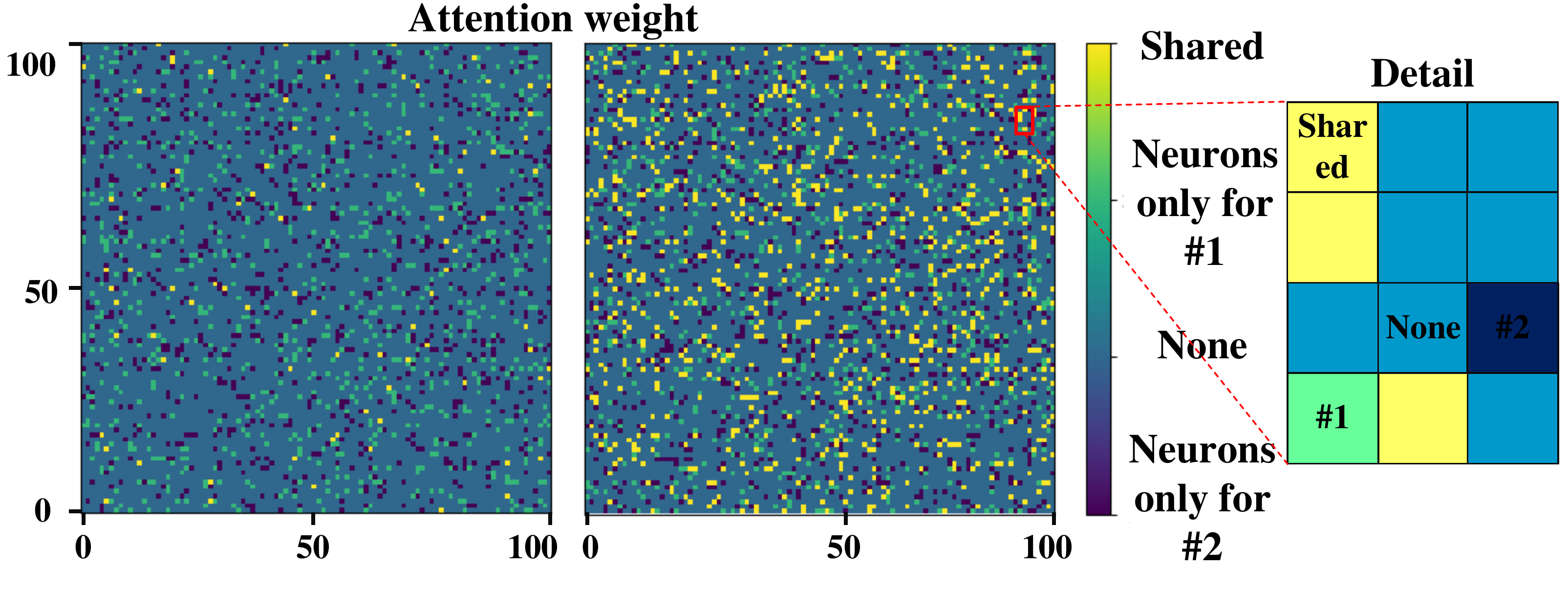}
  \put(5,-1){\small{(a) Disentanglement}}
  \put(41,-1){\small{(b) Similarity}}
  % \put(47,-1){\small{Float4}}
  \end{overpic}
  
  \caption{Disentanglement of concept neurons. (a) The intersection of concept neurons for two different subjects is sparsely distributed. (b) The concatenation of concept neurons for two different subjects is similar to the result of computing concept neurons from a joint loss of both subjects.}
  \label{fig:moti_iou}

\end{figure}

\begin{figure}[t]
  \centering
    \begin{overpic}[width=0.96\linewidth]{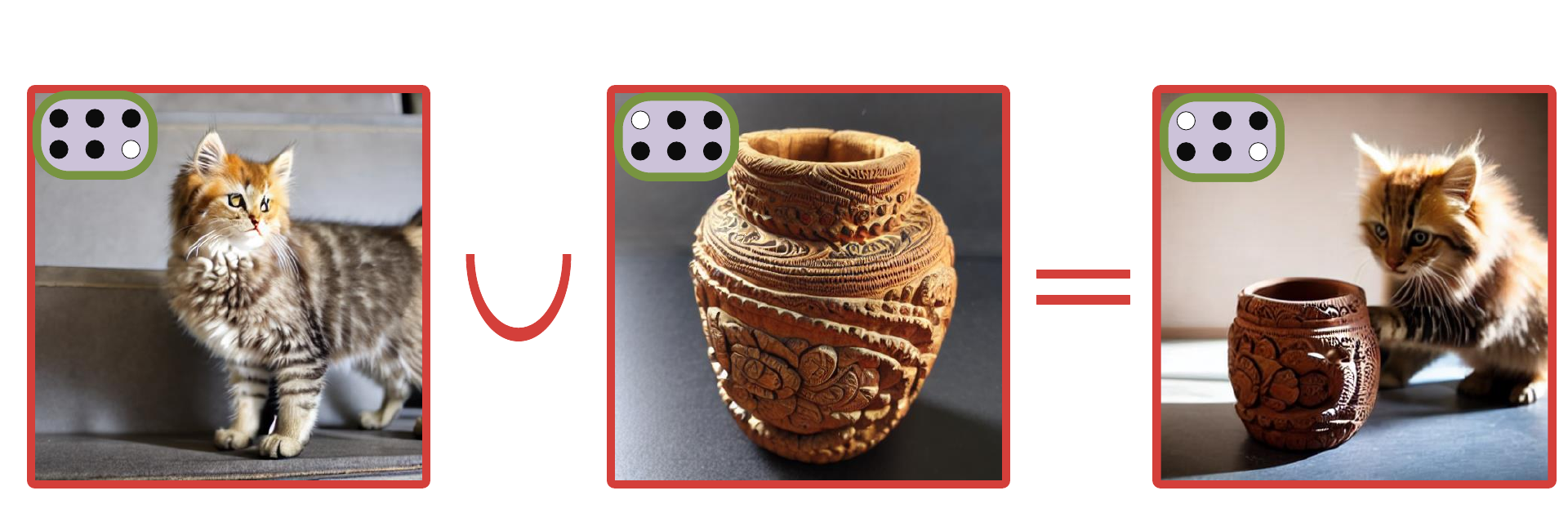}
  \put(74,18.5){\small{Neuron}}
  \put(4,-1.5){\small{Subject \#1.}}
  \put(43,-1.5){\small{Subject \#2.}}
  \put(77,-1.5){\small{Naive \#1+\#2.}}
  \put(4,30){\small{A $V_{1}*$ cat.}}
  \put(34,30){\small{A $V_{2}*$ wooden pot.}}
  \put(68,30){\small{\makecell{A $V_{1}*$ cat playing with \\ a $V_{2}*$ wooden pot.}}}
  \end{overpic}
  
  \caption{Additivity of concept neurons. Directly concatenating concept neurons of a subject cat and a subject wooden pot can vividly generate them both in the output under the direction of the text prompt. More results of additivity can be found in~\cref{sec:exps}.}
  \label{fig:moti_mask}
\vspace{-0.33cm}
\end{figure}
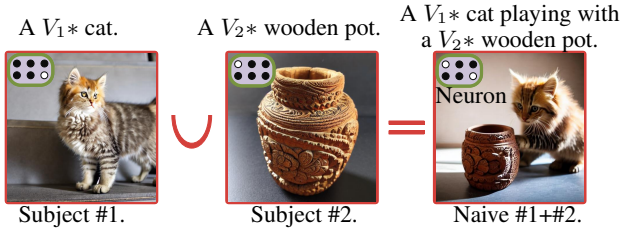

\myparagraph{Additivity of Concept Neurons for Multi-Subject Generation.}\label{sec:inte}
The concept neurons admit additivity. Direct joining concept neurons of multiple different subjects will yield concept neurons to generate the combination of them. This is the first time we manage to find an intrinsic affine semantic structure in the parameter space of diffusion models. In~\cref{fig:moti_mask}, we report the result of directly concatenating the concept neurons for a subject cat and wooden pot. After shutting the concatenated neurons, the diffusion model can immediately generate the two subjects together when accepting text prompts of the two corresponding identifiers as inputs. We will give more examples of the additivity for two and three concepts in~\cref{sec:exps}. Directly concatenating concept neurons can be an efficient method for customized multi-subject generations.

% \begin{figure*}[t]
%   \centering
%     \begin{overpic}[width=0.96\linewidth]{fig/moti_mask.pdf}
%   \put(1.5,13){\rotatebox{90}{\small{Pretrained model}}}
%   \put(34,13.5){\rotatebox{90}{\small{Concept addition}}}
%   \end{overpic}
%   \caption{Direct Addition of Concept.}
%   \label{fig:moti_mask}
%   \vspace{-0.1in}
% \end{figure*}

\subsection{Collaboratively Capturing Multiple Concepts}\label{sec:coll}
\begin{figure}[t]
    \centering
    \includegraphics[width=1.02\linewidth]{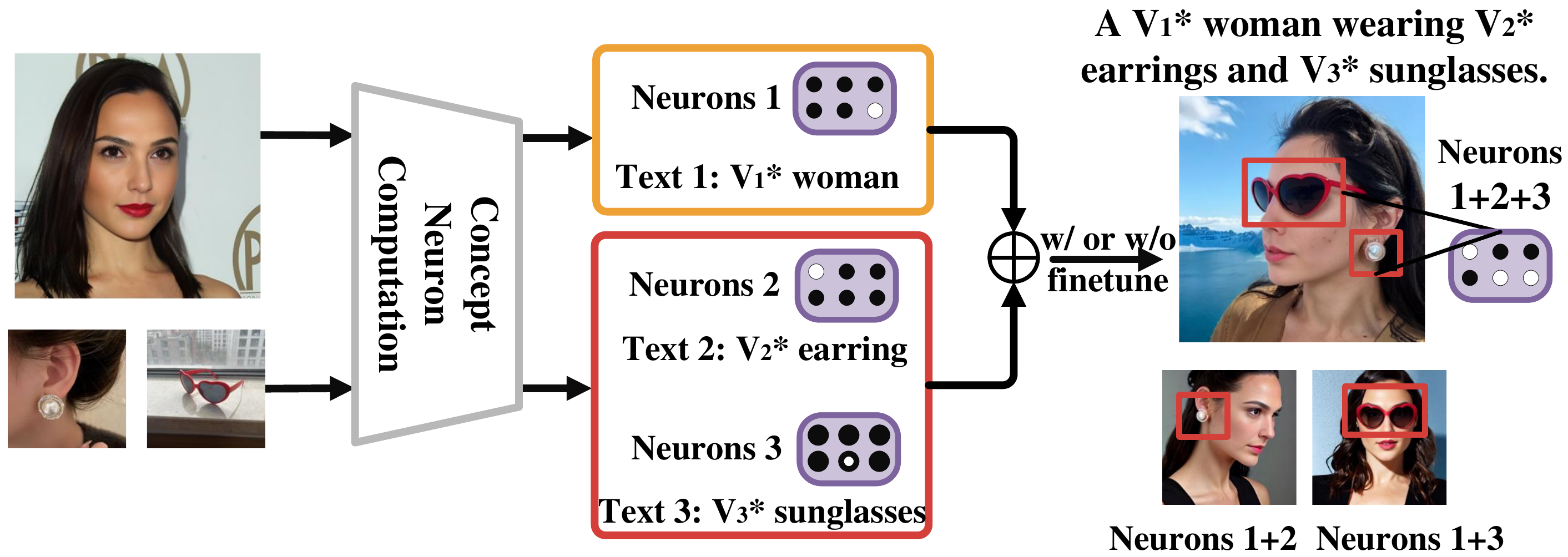}

    \caption{
        Illustration of collaborative capturing of multi-subject. Here we fine-tune the concatenation of concept neurons of multiple subjects to find a finer concept neuron mask.
    }

    \label{fig:sampling}
    \vspace{-0.2cm}
\end{figure}

For better generation quality and multi-subject generating capability, we can further fine-tune the concept neurons after concatenating. We calculate the sum of concept-implanting losses for all the involved subjects as a multi-concept-implanting loss. We then replace $\con$ in~\cref{algorithm:CN} with it and limit the computation in the concatenation of concept neurons for single subjects, as is shown in~\cref{fig:sampling}. This step will eliminate subtle conflicts in the concatenation of concept neurons due to inaccuracy in previous computations. The computed results can be more powerful in generating multi-subject. As we will show in~\cref{sec:exps}, this is the first work to generate up to four different diverse subjects contextualized in one image.

 Empirically we find the above pipeline slightly better than learning the concept neurons for multi-subject from scratch, \ieno, searching concept neurons in the whole K-V attention layers. This could be due to the increasing difficulty and instability of learning a loss landscape of complicated components. Besides, as we will show later, concept neurons enjoy good disentanglement; learning based on their concatenation could be stable and efficient.

\myparagraph{Disentanglement of Concept Neurons.} Concept neurons are well disentangled, which may be part of the reason for their additivity. \cref{fig:moti_iou} illustrates the concept neurons for the cat and wooden pot in~\cref{fig:moti_mask} (in layer ${\rm upblocks.2.attentions.1.transformerblocks.0.attn1.tov}$ of the StableDiffusion V1.4). We can find that the shared neurons between two independent concepts are very sparse, counting merely 2.42\% of the total concept neurons. We also report the differences between the concatenating two clusters of concept neurons and the result of learning from scratch. We can find the result of learning from scratch is close to the concatenation---they share 53.27\% neurons. Thus when only two subjects are involved, learning based on concatenation performs similarly to learning from scratch. When involving more subjects, the increasing complexity of loss function often spoils learning from scratch. Learning from concatenation, on the other hand, provides a much more comfortable starting point and reduces the overall difficulty of the task.

\subsection{Efficient Storage}
Previous customized generation methods~\cite{dreambooth,cosdiff}demand to save the parameters of the diffusion model in full digital accuracy. This can cost
considerably for large-scale applications in mobile devices.
While the concept neurons are sparse and binary, we only
need to record a small collection of indices for them. Those
indices can be stored with int instead of float data type,
thus further reducing the storage consumption. As we will
discuss in~\cref{sec:exps}, Cones requires no more than 10\% memory of previous customized generation methods.

% \subsection{Efficient Storage Method}\label{sec:store}
% The attention layer of k, v is a matrix $R^{h\times w}$, the concept neurons are very environmental-friendly to save as we only need to store a sparse cluster of {\rm int} index instead of dense {\rm float32} values in the matrix. In order to save space, we store the stored parameters in bits instead of bytes. Specifically, we calculate how many bits need to be set to satisfy the maximum index of the current matrix
% \begin{equation}
%     \lceil x \rceil=\log_2^{h*w}
% \end{equation}
% where $\lceil x \rceil$ represents the bits required to store the index in the current matrix. Assume that the parameters we need to store in the current matrix are $a$, which the mean of $a$ in all k, v matrices is $0.5\%-2\%$ (changing according to different concepts). In addition, we use $8$ bits to represent the current value of $\lceil x \rceil$. Using this encoding method can prevent the waste of space caused by using the int32 type for storage. In general, we use float32 to store all weights in the matrix, and the bit-level storage space used is $h*w*32$. Therefore, the storage space we need compared to the original is:
% \begin{equation}
%     y=\frac{\lceil x \rceil *a*h*w+8}{32*h*w}
% \end{equation}

% \begin{figure}[t]
%     \centering
%     \includegraphics[width=1.02\linewidth]{fig/denoising.pdf}
%     \caption{ 
%         Mask Conditioning Denoising.
%     }
%      \vspace{-.15in}
%     \label{fig:finetune}
% \end{figure}

%% file: section/4.exps.tex
\section{Experiments}\label{sec:exps}
\subsection{Implementation and Experiment Details }

\begin{figure*}[t]
  \centering
    \begin{overpic}[width=0.96\linewidth]{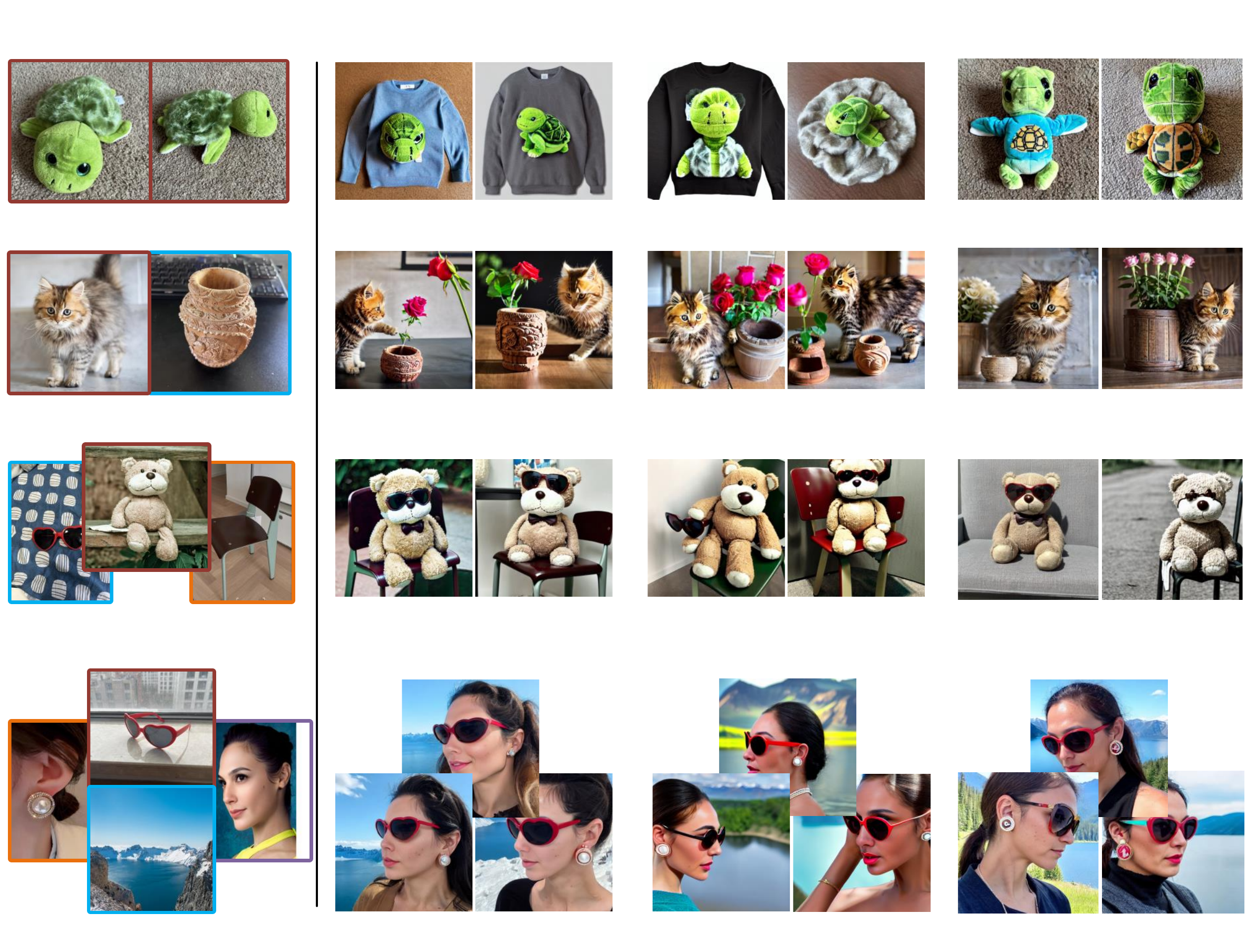}
  \put(6,58){\small{(a) One subject.}}
  \put(6,42.8){\small{(b) Two subjects.}}
  \put(6,26){\small{(c) Three subjects.}}
  \put(6,0.5){\small{(d) Four subjects.}}
  \put(6,72){\small{Reference images}}
  \put(34,72){\small{Cones (Ours)}}
  \put(56.8,72){\small{Custom Diffusion}}
  \put(82,72){\small{Dreambooth}}
  \put(42,58){\small{A $V_{1}*$ tortoise plushy printed on a sweater.}}
  \put(36,41.5){\small{\makecell{A $V_{1}*$ cat is playing with $V_{2}*$ wooden pot on a table, \\ and a rose in the $V_{2}*$ wooden pot.}}}
  \put(39,25){\small{\makecell{A $V_{1}*$ chair with a $V_{2}*$ teddybear sitting on it, \\ and the $V_{2}*$ teddybear is wearing $V_{3}*$sunglasses.}}}
  \put(36,0.2){\small{\makecell{A $V_{1}*$ woman is wearing $V_{2}*$ earrings on her ear, \\ and  $V_{3}*$ 
  sunglasses on her face, with $V_{4}*$ lake in the background.}}}
  \end{overpic}
  \caption{Comparison of multi-subject generation ability. First row: compared with other methods, ours can better generate the ``sweater'' in the prompt. Second row: Our method better reflects the semantics of ``playing'', while Dreambooth loses the details of the wooden pot. Third row: our generated images have a higher visual similarity with the target subject, and better semantics alignment with ``sitting'' and ``wearing''. Dreambooth fails to generate ``chair''. Fourth row: Cones (Ours) maintains high visual similarity for all subjects.
  }
  \label{fig:visual_jonit}
 
\end{figure*}
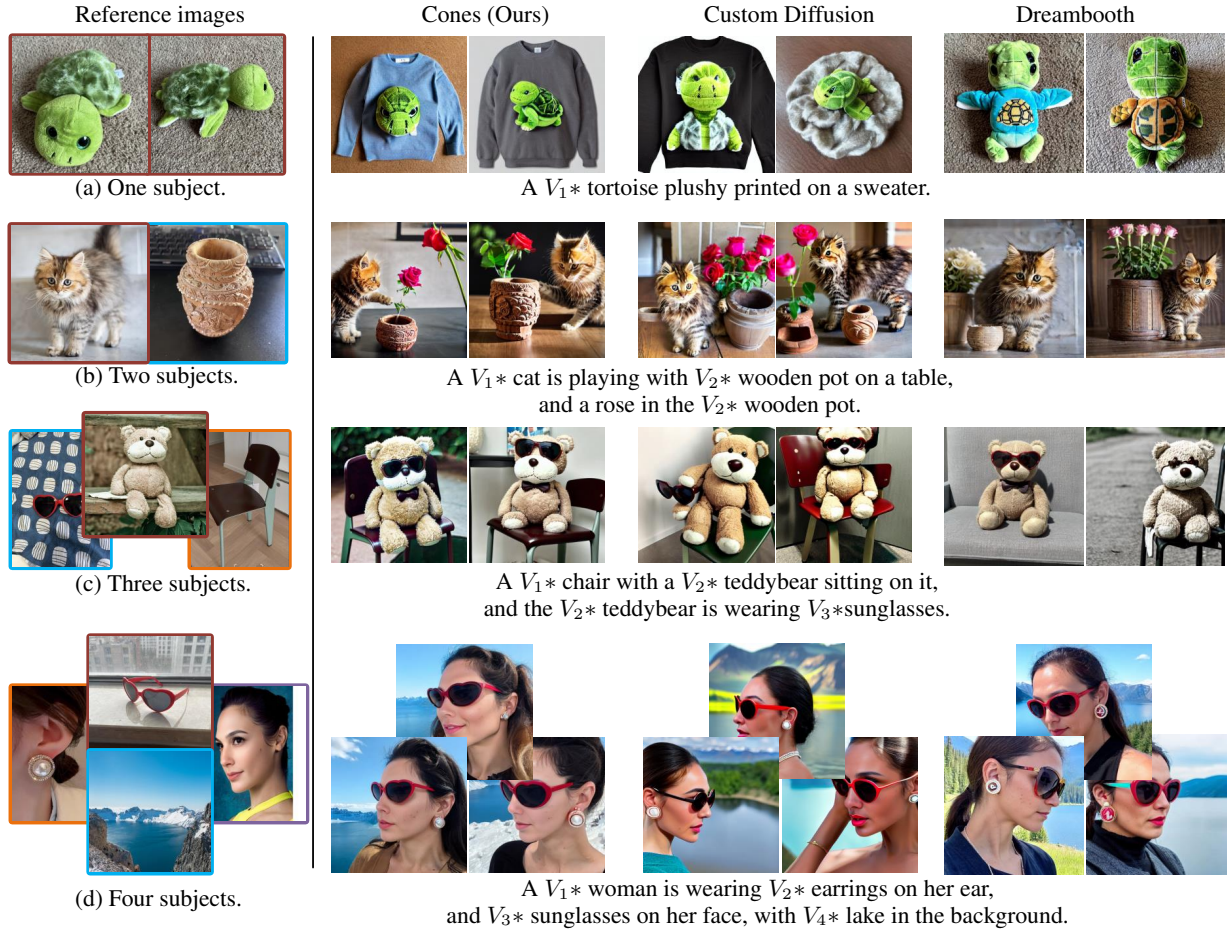

\myparagraph{Evaluation Metrics, Datasets, and Implementation Details.}
We evaluate Cones with two famous metrics for customized generation proposed in Textual Inversion~\cite{gal2022image}. (1) Image alignment, which measures the visual similarity between the generated images and the target concept. Specifically, we use the CLIP~\cite{radford2021learning} model (ViT-L/14, consistent with the text encoder in Stable Diffusion V1.4) to calculate the CLIP-space cosine-similarity between the generated images and the target subject. For multi-subject generation, we calculate the image alignment of the generated images and each target subject separately and finally calculate the mean value. (2) Text alignment, which evaluates the ability of Cones to edit the target subjects with text prompts. To this end, we use a variety of prompts with different settings to generate images, including modifying the background, style, and attributes. We calculate the average CLIP-space embedding of the generated images and compute their cosine similarity with the CLIP-space embedding of the textual prompts, where we omit the text identifier in textual prompts. All images used in the paper are downloaded from anonymous e-commerce websites or Unsplash, like the dataset of Custom Diffusion~\cite{cosdiff}. Implementation details are reported in~\cref{sec:exp_sets}.

\myparagraph{Competing Methods.}
To evaluate our generation quality and multi-subject generation capability, we compare Cones with three competitors. They are Dreambooth~\cite{dreambooth} that fine-tunes all parameters in the diffusion model; Text Inversion~\cite{gal2022image} that adds a new token for each new concept and only updates the new token embedding during fine-tuning; and Custom Diffusion~\cite{cosdiff} that optimizes the newly added token embedding in text encoder and a few parameters in diffusion model, namely the key and value mapping from text to latent features in the cross-attention~\cite{yu2022scaling, vaswani2017attention} layers.

\subsection{Qualitative Evaluation}
To demonstrate the effectiveness of Cones, we conduct experiments on authentic images of diverse categories, including objects, backgrounds, portraits, \etcno. As shown in \cref{fig:visual_jonit}, we show the results of generating several subjects in the same scene for the following four settings: (1) single subject: tortoise plushy, (2) two subjects: cat + wooden pot, (3) three subjects: chair + teddybear + sunglasses, (4) four subjects: woman + earrings + sunglasses + lake. For each method and subject setting, we sample 20 output images using 20 random seeds. We then select the best two of the 20 images as candidates for comparison. We omit Textual Inversion as it performs much less competitively. We thus put the generated results of Textual Inversion in~\cref{fig:supp_inversion}. As shown in \cref{fig:visual_jonit} (a), Cones is on-par with two other methods for visual similarity, but Cones has higher alignment with the input prompt than other methods for the single subject. As more subjects are composed together, Cones can generate images with good visual accuracy for all subjects. In contrast, the other two methods will make some subjects disappear or become less similar to the reference images. 

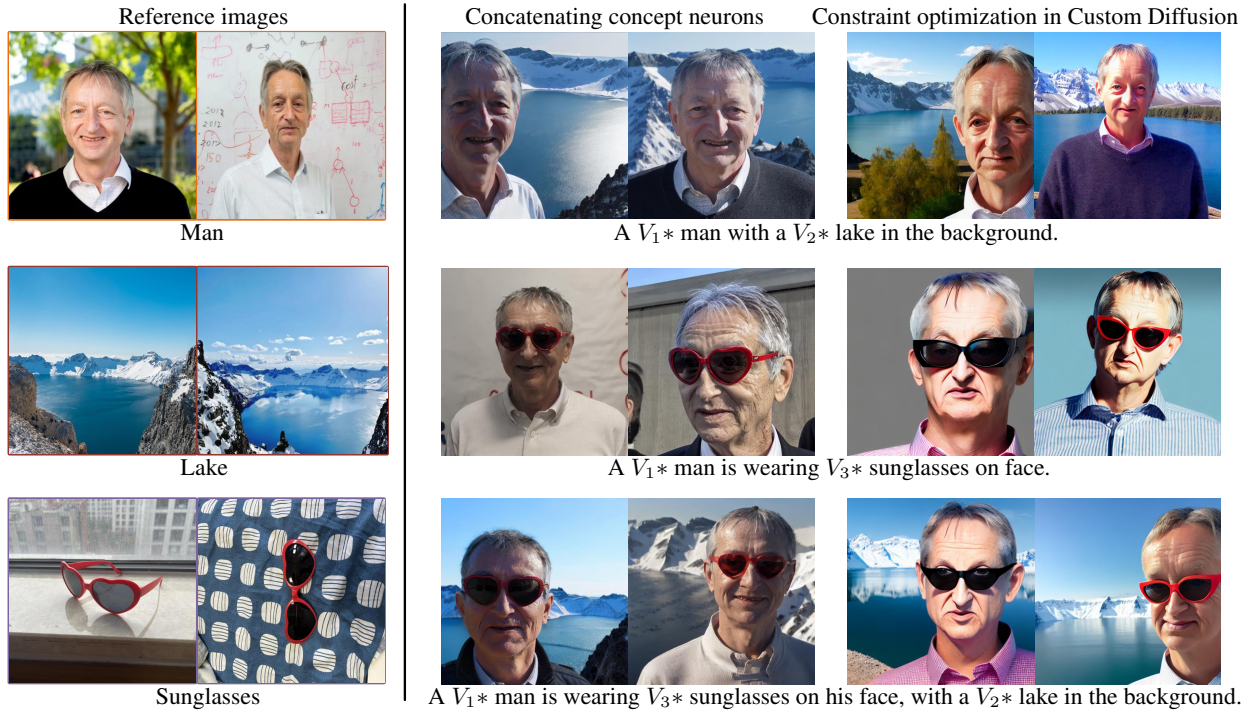
\begin{figure*}[t]
  \centering
    \begin{overpic}[width=0.96\linewidth]{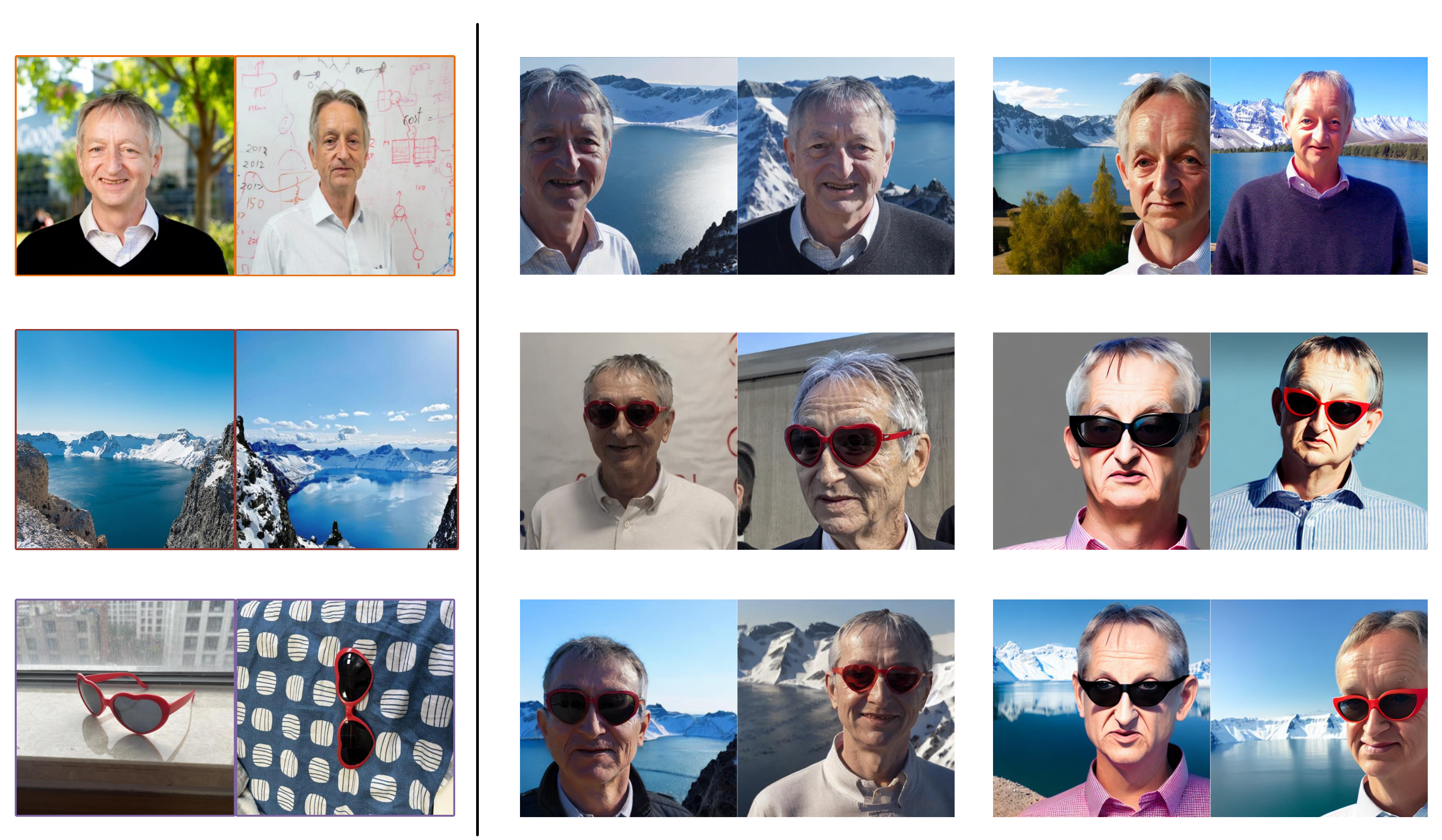}
  \put(15,37.5){\small{Man}}
  \put(15,18.5){\small{Lake}}
  \put(13,0){\small{Sunglasses}}
  \put(10,55){\small{Reference images}}
  \put(38,55){\small{Concatenating concept neurons}}
  \put(66,55){\small{Constraint optimization in Custom Diffusion}}
  \put(50,37.5){\small{A $V_{1}*$ man with a $V_{2}*$ lake in the background.}}
  \put(49.5,18.5){\small{A $V_{1}*$ man is wearing $V_{3}*$ sunglasses on face.}}
  \put(35,0){\small{A $V_{1}*$ man is wearing $V_{3}*$ sunglasses on his face, with a $V_{2}*$ lake in the background.}}
  \end{overpic}
  \caption{Comparison of tuning-free subject generation methods. For Cones, we concatenate concept neurons of multiple subjects directly. For Custom Diffusion, we use the ``constraint optimization'' method of it to composite multiple subjects.}
  \label{fig:mask_concat}

\end{figure*}

\myparagraph{Tuning-Free Comparison.}
By concatenating two clusters of concept neurons, we can realize the composition of the two concepts without further fine-tuning. While Custom Diffusion also provides a tuning-free method to composite multiple subjects (the ``constraint optimization'' method), we compare our concatenation of concept neurons with it in \cref{fig:mask_concat}. It is easy to see that the concatenation of concept neurons significantly outperforms the tuning-free composition of the Custom Diffusion in visual quality and subject-generation accuracy.

\subsection{Quantitative Evaluation and User Study}
\myparagraph{Quantitative Evaluation.}
We evaluate 20 prompts for each concept group and generate 50 images per prompt. For the multi-subject generation, we represent the image alignment as the mean of the visual similarity between the generated image and all target concepts. As shown in \cref{tbl:results}, For the single subject generation, our visual accuracy is comparable to Custom Diffusion, slightly lower than Dreambooth. Yet Cones has higher text alignment, which means Cones captures those subjects better and is more faithful to the prompt itself. When the number of involved subjects increases, Cones outperforms the competitors in all metrics.

\begin{table}[!t]
\centering
\setlength{\tabcolsep}{5pt}
\resizebox{\linewidth}{!}{
\begin{tabular}{ll c c }
\toprule
& \textbf{Method}
& \shortstack[c]{\textbf{Text-alignment} } 
& \shortstack[c]{\textbf{Image-alignment} }
 \\
\midrule
\multirow{4}{*}{\shortstack[c]{\textbf{Single-}\\ \textbf{Subject} } }  & Textual Inversion    & 0.312 & \textbf{0.744}  \\
& DreamBooth & 0.344   & 0.731 \\

& Custom Diffusion   & 0.352 & 0.722 \\
& Cones (Ours)    & \textbf{0.361} & 0.725 \\
\midrule

\multirow{4}{*}{\shortstack[c]{\textbf{Two-}\\ \textbf{Subjects} } } 
&   Textual Inversion   & 0.264 & 0.630 \\
& DreamBooth & 0.283 & 0.673 \\
& Custom Diffusion & 0.314 & 0.685  \\
& Cones (Ours)  & \textbf{0.337} & \textbf{0.698}\\
\midrule

\multirow{4}{*}{\shortstack[c]{\textbf{Three-}\\ \textbf{Subjects} } } 
&   Textual Inversion   & 0.223 & 0.584 \\
& DreamBooth & 0.263 & 0.631 \\
& Custom Diffusion & 0.289 & 0.669  \\
& Cones (Ours)  & \textbf{0.301} & \textbf{0.685} \\
\midrule
\multirow{4}{*}{\shortstack[c]{\textbf{Four-}\\ \textbf{Subjects} } } 
&   Textual Inversion   & 0.219 & 0.553 \\
& DreamBooth & 0.238 & 0.597 \\
& Custom Diffusion & 0.269 & 0.632  \\
& Cones (Ours)  & \textbf{0.285} & \textbf{0.653} \\
\bottomrule

\end{tabular}
}

\caption{Quantitative comparisons. Cones performs the best except for image alignment in the single subject case. This could be due to that the image alignment metric is easy to overfit as is pointed out in Custom Diffusion~\cite{cosdiff}. DreamBooth and Textual Inversion employ plenty of parameters in the learning, while Cones only involves the deactivation of a few parameters.}

\label{tbl:results}

\end{table}

\myparagraph{User Study.}
We conduct a user study to further evaluate Cones. Two questions are designed to measure the image alignment and text alignment of all the methods. For text alignment, we ask the users ``which image is most consistent with the textual description in the prompt''. For image alignment, we ask the user ``Which image is the most similar to the provided reference images''. We hire 50 annotators to answer each of those questions. Details of how we conduct user study can be found in appendix. As shown in~\cref{tbl:human_eval}, Cones performs the best in all cases, earning the most votes, except for image alignment in the single subject generation. 
\begin{table}[!t]
\centering
\setlength{\tabcolsep}{5pt}
\resizebox{\linewidth}{!}{
\begin{tabular}{l c c }
\toprule
\textbf{Method}
& \textbf{Storage}
& \textbf{Sparsity}
 \\
\midrule
\textbf{Dreambooth}    & 3.3GB & -- \\
\textbf{Custom Diffusion} & 72MB   & --  \\
\textbf{Ours (single subject)}   & 1.43MB$\pm$ 0.34MB & 1.32\% $\pm$ 0.29$\%$ \\
\textbf{Ours (two subjects)}   & 3.41MB $\pm$ 0.56MB& 2.43\% $\pm$ 0.44$\%$ \\
\textbf{Ours (three subjects)}   & 4.96MB $\pm$ 0.70MB & 4.54\% $\pm$ 0.59$\%$ \\
\textbf{Ours (four subjects)}   & 7.75MB$\pm$ 0.56MB &  7.01\% $\pm$ 0.26$\%$ \\
\bottomrule

\end{tabular}
}

\caption{Storage cost and sparsity of concept neurons. As the number of target subjects increases, we need to store more indexes of concept neurons. We save more than 90\% of the storage space compared with Custom Diffusion,
}
\label{tbl:ablation}

\end{table}

\myparagraph{Sparsity and Storage.}
Thanks to the sparsity of concept neurons, we only need to record a small collection of indexes for them in attention layers. Those indices can be stored with int instead of float data type. As shown in~\cref{tbl:ablation}, we show the storage required by Cones for multi-subject generation, and the sparsity of the corresponding concept neurons. Here sparsity means the percentage of concept neurons in all the neurons of the attention layers. We can find that Cones costs much less storage compared with the competitors.

%% file: section/5.conclusion.tex
\section{Conclusion}\label{sec:conclus}
This paper reveals concept neurons in the parameter space of diffusion models. We find that for a given subject, there is a small cluster of concept neurons that dominate the generation of this subject. Shutting them will yield renditions of the given subject in different contexts based on the text prompts. Concatenating them for different subjects can generate all the subjects in the results. Further fine-tuning can enhance the multi-subject generation capability, which is the first to manage to generate up to four different subjects in one image. Comparison with state-of-the-art competitors demonstrates the superiority of using concept neurons in visual quality, semantic alignment, multi-subject generation capability, and storage consumption.

%% file: section/6.ref.tex
\bibliography{ref}
\bibliographystyle{icml2023}

%% file: section/7.appendix.tex
\renewcommand\thefigure{A\arabic{figure}}
\renewcommand\thetable{A\arabic{table}}  
\renewcommand\theequation{A\arabic{equation}}
\renewcommand\thealgorithm{A\arabic{algorithm}}

\onecolumn
\appendix

\section*{Appendix}

%\tableofcontents
\input{section/7.1.proofs.tex}
\input{section/7.2.exp_sets.tex}
\input{section/7.3.more_results.tex}

\begin{figure}[t]
  \centering
    \begin{overpic}[width=0.96\linewidth]{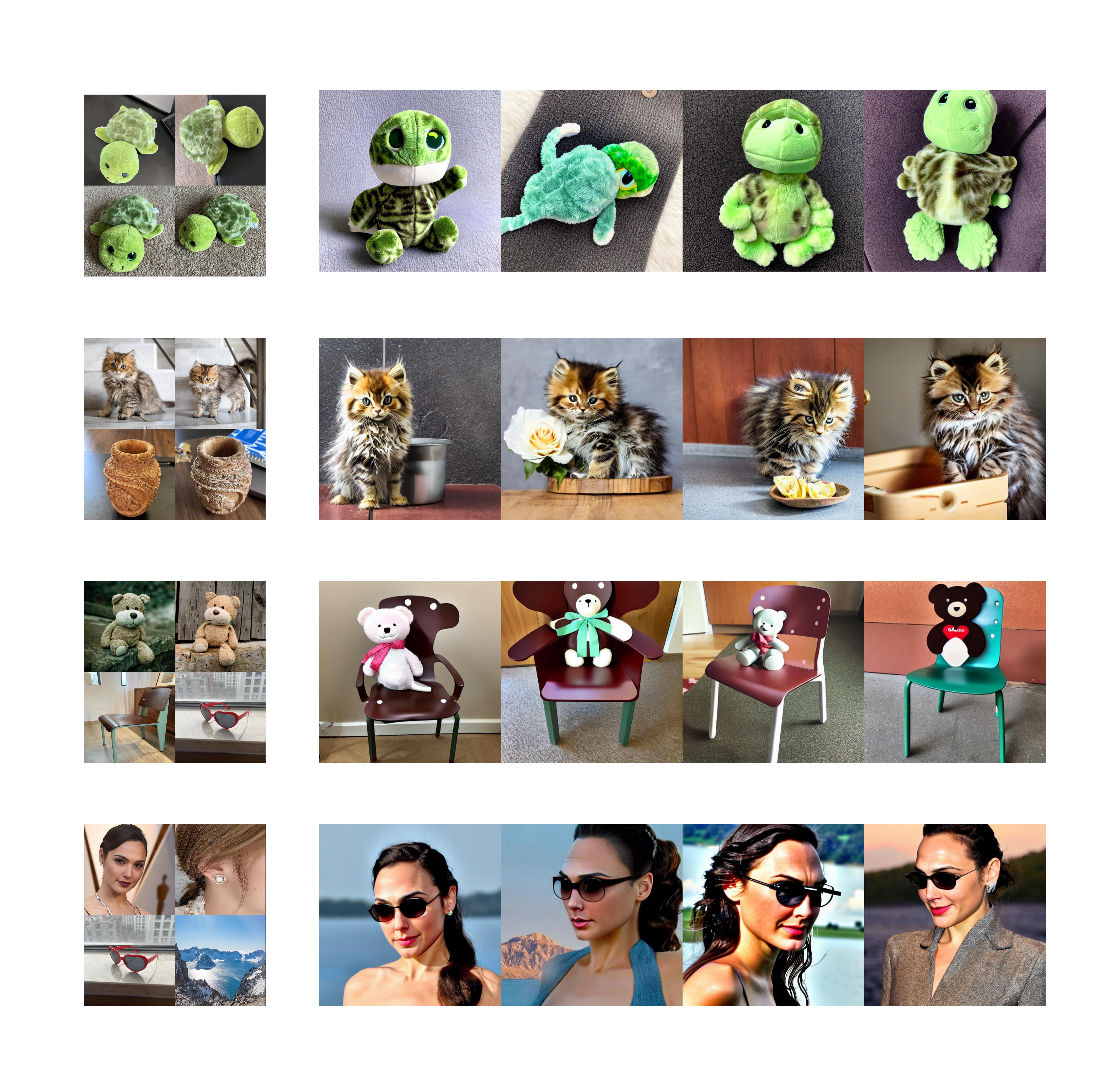}
  \put(9.5,69.5){\small{(a) One subject.}}
  \put(9.5,47.5){\small{(b) Two subjects.}}
  \put(9.5,25.5){\small{(c) Three subjects.}}
  \put(9.5,4){\small{(d) Four subjects.}}
  \put(9.5,90.5){\small{Reference images}}
  \put(55.5,90.5){\small{Textual Inversion}}
  \put(54,70){\small{\makecell{A $V_{1}*$ tortoise plushy \\ printed on a sweater.}}}
  \put(51,48){\small{\makecell{A $V_{1}*$ cat is playing with \\ $V_{2}*$ wooden pot on a table.}}}
  \put(47,26){\small{\makecell{A $V_{1}*$ teddybear  wearing $V_{2}*$ \\sunglasses is sitting  on the $V_{3}*$ chair.}}}
  \put(37,4.5){\small{\makecell{A $V_{1}*$ woman is wearing $V_{2}*$ earrings on her ear, \\ and  $V_{3}*$ 
  sunglasses on her face, with $V_{4}*$ lake in the background.}}}
  \end{overpic}
  \caption{Multi-subject generation using Textual Inversion. We observe that Textual Inversion struggles with the composition of multiple subjects.}
  \label{fig:supp_inversion}
  \vspace{-0.1in}
\end{figure}

\begin{figure}[t]
  \centering
    \begin{overpic}[width=1.08\linewidth]{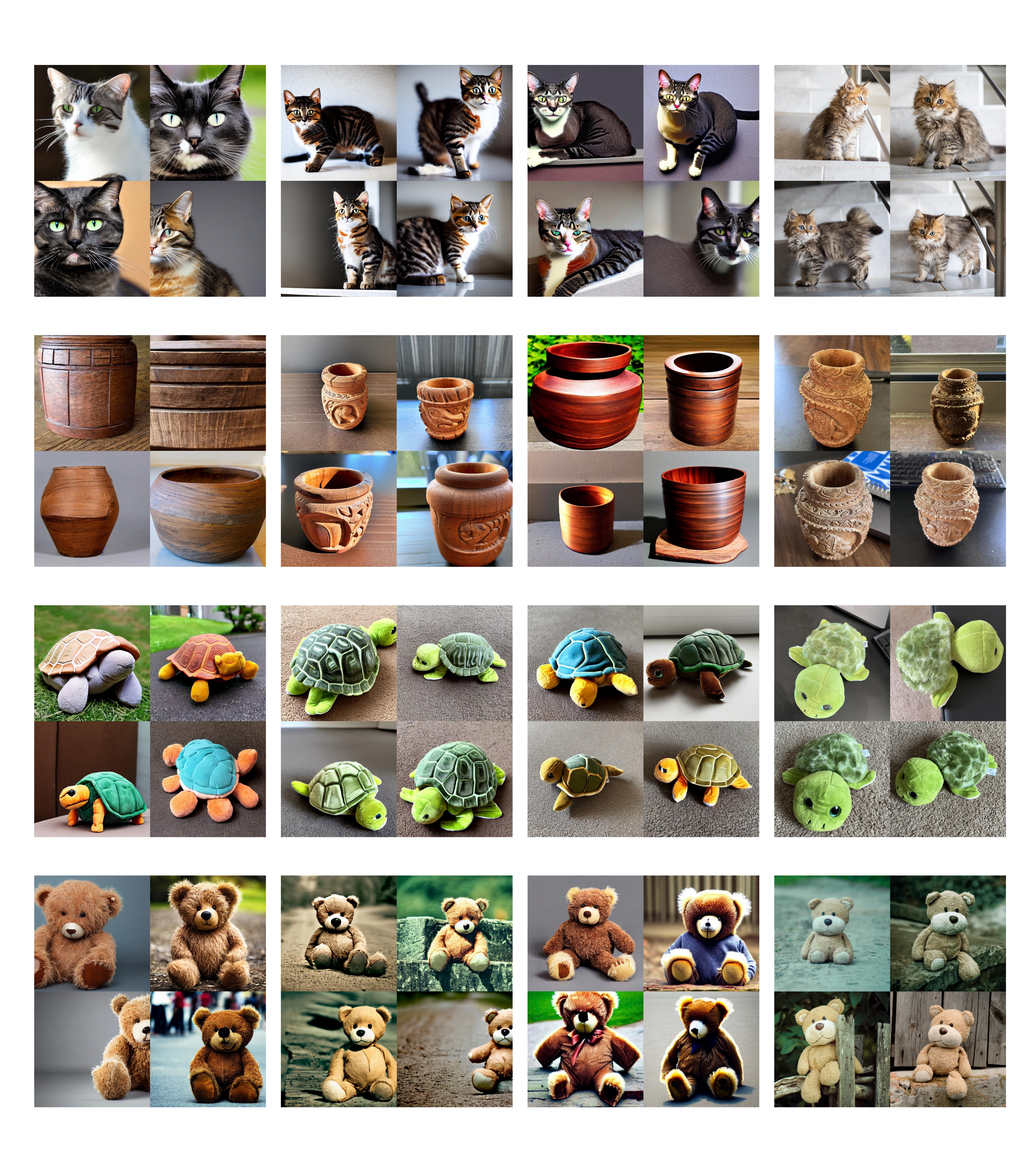}
  \put(5.8,95){{Pretrained Model}}
  \put(27,95){{Custom Diffusion}}
  \put(52.5,95){{Cones (Ours)}}
  \put(70,95){{Target Images}}
  \put(38,73){{Photo of a cat}}
  \put(36.5,50){{Photo of a wooden pot}}
  \put(36,27){{Photo of a tortoise plushy}}
  \put(36.5,3.5){{Photo of a teddybear}}
  
  \end{overpic}
  \caption{Overfitting on the training prompt template. The fourth column corresponds to the training data, and it can be seen that even without the text identifier, the generations of Custom Diffusion still retain some characteristics of the target images.The generations of Cones after finetuning has more diversity similar to the pretrained model.}
  \label{fig:overfitting}
  \vspace{-0.1in}
\end{figure}

\begin{figure}[t]
  \centering
    \begin{overpic}[width=0.96\linewidth]{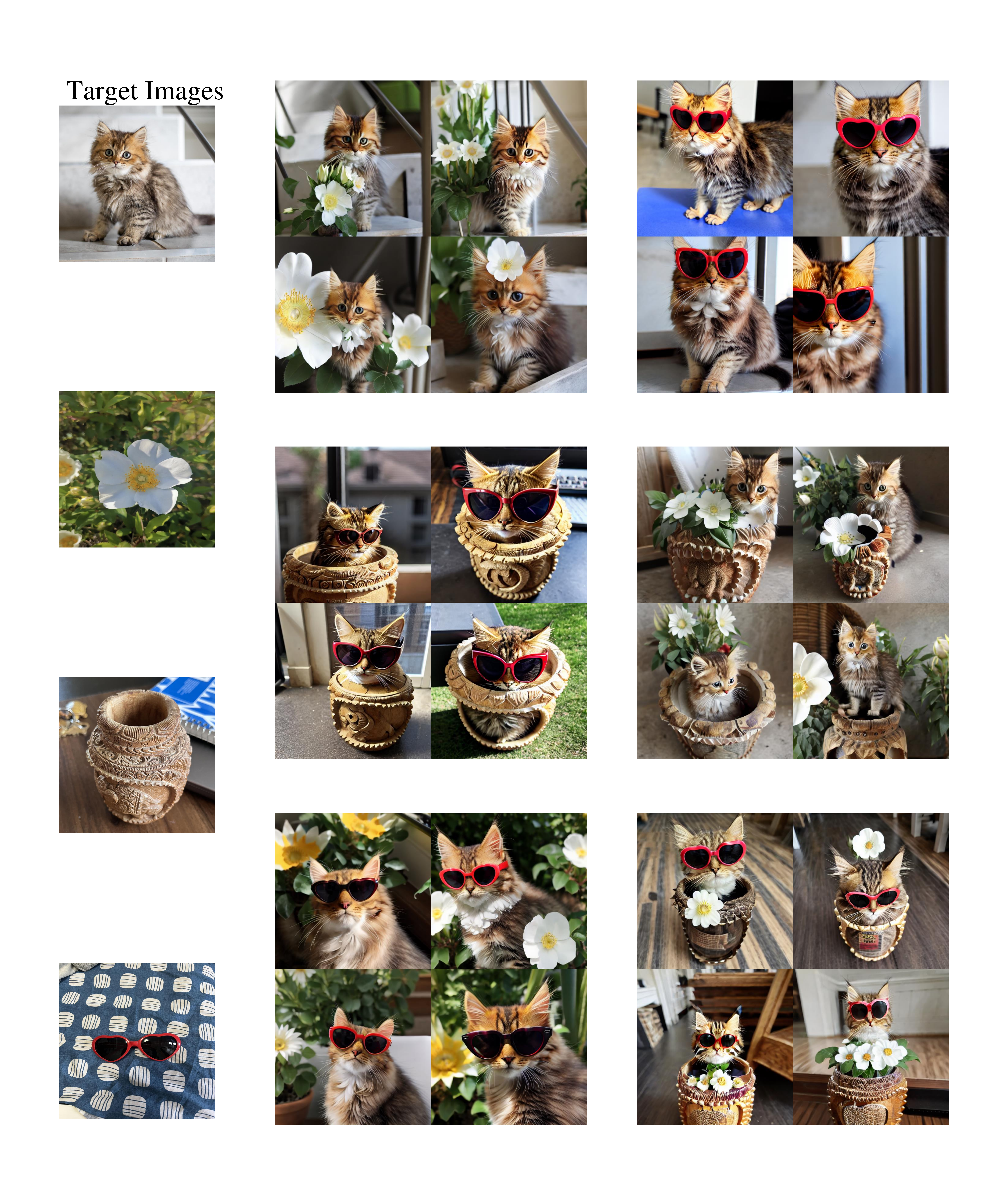}
  \put(25.2,64.5){{ $V_{1}*$ flowers next to a $V_{2}*$ cat.}}
  \put(55.6,64.5){{ A $V_{2}*$ cat with $V_{3}*$ sunglasses.}}
  \put(21,34.1){{\makecell{ A $V_{4}*$ barrel with a  $V_{2}*$ cat sitting inside it, \\ which is wearing $V_{3}*$ sunglasses.}}}
  \put(55.6,34.1){{\makecell{$V_{1}*$ flowers in a $V_{4}*$  barrel \\ and a $V_{2}*$ cat is playing with it.}}}
  \put(23.5,3.6){{\makecell{$V_{1}*$ flowers next to a $V_{2}*$ cat,\\ which is wearing $V_{3}*$ sunglasses.}}}
  \put(52.1,3.6){{\makecell{A $V_{4}*$ barrel decorated with $V_{1}*$ flowers, \\ and a $V_{2}*$ cat wearing $V_{3}*$ sunglasses.}}}
  \end{overpic}
  \caption{More results of multi-subject generation}
  \label{fig:more_multi}
  \vspace{-0.1in}
\end{figure}

%% file: section/7.1.proofs.tex
\section{Proof}\label{sec:proof}
\subsection{Proof to~\cref{th}}
This is easy to see from~\cref{eq:scale,eq:taylor,eq:equiv}.

\newcommand{\vga}{\bm{\gamma}}
\newcommand{\vxi}{\bm{\xi}}

\subsection{Further Acceleration of~\cref{eq:main}}
Let $\gamma=\xi\theta$ and $\vga=\vxi\odot\vtheta$. Using $\vga$ to replace $\vtheta$ as the parameter of $\diff$, and independent variable of $\con$, then by Newton-Leibniz law of calculus, we have
\begin{gather}
    \frac{\partial\con(\vga)}{\partial\xi}=\frac{\partial\con(\gamma)}{\partial\gamma}\frac{\partial\gamma}{\partial\xi}=\theta\frac{\partial\con(\gamma)}{\partial\gamma},\\
    \nabla_{\vxi}\con(\vxi\odot\vtheta)=\vtheta\odot\nabla_{\vga}\con(\vga)=\vtheta\odot\nabla_{\vtheta}\con(\vtheta)|_{\vtheta=\vga}.
\end{gather}
Thus it is easy to see, the gradient descent over function
\begin{equation}
    \con(\vxi\odot\vtheta)
\end{equation}
will yield update rule
\begin{gather}
    \vxi^{k+1}=\vxi^k-\beta\nabla_{\vxi}\con(\vxi\odot\vtheta)=\vxi^k-\beta\vtheta\odot\nabla_{\vga}\con(\vga^k),\\
    \vga^{k+1}=\vxi^{k+1}\odot\vtheta=\vxi^k\odot\vtheta-\beta\vtheta^2\odot\nabla_{\vga}\con(\vga^k)=\vga^k-\beta\vtheta^2\odot\nabla_{\vga}\con(\vga^k),
\end{gather}
where $\vtheta^2=\vtheta\odot\vtheta$.
When learning rate $\beta$ is small, $\vxi$ is initialized as $\bm{1}$, and iteration step $k$ is not large, $\vga^k\approx\vtheta$, thus
\begin{equation}
    \vga^{k+1}\approx\vga^k-\beta(\vga^k)^2\odot\nabla_{\vga}\con(\vga^k)=\vga^k\odot(\bm{1}-\beta\vga^k\odot\nabla_{\vga}\con(\vga^k)).
\end{equation}
Note that this is actually~\cref{eq:main} and our sampling rule in~\cref{eq:sampling}. 

Thus, we can accelerate the computation of neuron concepts by setting $\vga=\vxi\odot\vtheta$, initializing $\xi$ at $\bm{1}$, and conducting gradient descent on $\con(\vxi\odot\vtheta)$ with learning rate $\beta$ and optimization variable $\vxi$. When
\begin{equation}
    \vxi^K\approx 1-\beta(\vga^1\odot\nabla_{\vga}\con(\vga^1)+\vga^k\odot\nabla_{\vga}\con(\vga^k)),
\end{equation}
we have
\begin{equation}
    \mM_p=\frac{1}{\beta}(\bm{1}-\vxi^K).
\end{equation}
So the concept neuron mask can be computed as
\begin{equation}
    \mM=\bm{1}-(\mM_p>\tau)=\bm{1}-(\vxi^K<\bm{1}-\beta\tau).
\end{equation}
Thus we can use~\cref{algorithm:CNP} to compute the concept neuron mask in practice.

\begin{algorithm}[b]
	\caption{Accelerated Computation of Concept Neuron Mask}
	\label{algorithm:CNP}
	\begin{algorithmic}
		\STATE {\bfseries Input:} Concept-implanting loss function $\con$ with parameter $\vtheta\in\sR^n$ replaced by $\vtheta=\vxi\odot\vtheta$, training step $K$, learning rate $0<\rho\ll1$ and $\tau>0$.

		\STATE {\bfseries Execute:} $K$-step gradient descent updates to variable $\vxi$ with loss function $\con(\vxi\odot\vtheta)$ and learning rate $\rho$.
    \STATE {\bfseries Compute: } $\mM_p=\frac{1}{\rho}(\bm{1}-\vxi)$.
    \STATE{\bfseries Set:} $\mM=\bm{1}-(\mM_p>\tau)$.
	\STATE {\bfseries Output:} Binary concept neuron mask $\mM$ to indicate whether each neuron is a concept neuron, 1 for not and 0 for is.
	\end{algorithmic} 
\end{algorithm}

%% file: section/7.2.exp_sets.tex
\section{Experiment Setups}\label{sec:exp_sets}
We supplement the experimental Setups of each method in this section. Consistent with the Custom Diffusion~\cite{cosdiff} setting, we use Stable Diffusion V1.4 as the pretrained model. For a fair comparison, we use 50 steps of DPM-Solver~\cite{lu2022dpm} sampler with a scale 7.5 for all above methods. All experiments are conducted using an A-100 GPU. For the three methods, except for Textual Inversion, training steps increase linearly as the number of involved subjects increases, and we initialize the identifier with the same rare occurring token as in Custom Diffusion.
\subsection{Textual Inversion}
We train with the recommended\footnote{https://github.com/rinongal/textual\_inversion}
batch size of 4, a learning rate of 0.005 (scaled by batch
size for an effective learning rate of 0.02) for 5,000 steps. The new token embedding is initialized with the category name. When some categories require multiple tokens to represent, we choose to use an approximation word to summarize the multiple tokens, such as replacing "wooden pot" with "pot". 
\subsection{Dreambooth}
We use the third-party implementation of huggingface~\cite{von-platen-etal-2022-diffusers} for Dreambooth\footnote{https://github.com/huggingface/diffusers/tree/main/examples/dreambooth}. Training is with a batch size of 1, learning rate $5 \times 10^{-6}$, and training steps of 800.
\subsection{Custom Diffusion}
We use the official implementation\footnote{https://github.com/adobe-research/custom-diffusion} for Custom Diffusion, which is consistent with paper, \textit{i.e.}, the batch size is set to 4, training steps is set to 600 and the basic learning rate is $10^{-5}$ and scaled by batch size for an effective learning rate of $4 \times 10^{-5}$. 
\subsection{\methodbar(Ours)}
Our experiments are conducted on an A-100 GPU with a batch size of 2. We use~\cref{algorithm:CNP} to find the concept neurons. The base learning rate is set to $3 \times 10^{-5}$. we further scale the base learning rate of $6 \times 10^{-5}$ by the number of GPUs and the batch size.For the single-subject generation, the base learning rate is set to $2 \times 10^{-5}$, which can get better results. We train 1,000 steps for a single subject.

\subsection{User Study}
For one- to four-subject generation tasks, we design three different subject combinations for each task. This will yield 12 subject combinations in total. For each subject combination, we design four different text prompts to generate images. Each text prompt will be combined with 50 random seeds to generate 50 outputs. The best 2 are selected to represent the result of the corresponding text prompt. We conduct this procedure to all four methods, which results in 48 image octuples. Each octuple contains two best images generated by each method with each text prompt and subject combination. The results of user study can be found in~\cref{tbl:human_eval}.

\begin{table*}[!t]
\centering
\setlength{\tabcolsep}{6pt}
\resizebox{\linewidth}{!}{
\begin{tabular}{@{\extracolsep{4pt}}ll c cc cc cc@{} }
\toprule

&  \multicolumn{2}{@{} c}{\textbf{Textual Inversion}} 
&  \multicolumn{2}{@{} c}{\textbf{DreamBooth}} 
&  \multicolumn{2}{@{} c}{\textbf{Custom Diffusion}} 
&  \multicolumn{2}{@{} c}{\textbf{Ours}}
\\

\cmidrule{2-3} \cmidrule{4-5} \cmidrule{6-7} \cmidrule{8-9}
 &  \multirow{2}{*}{\shortstack[c]{Text\\ Alignment }}
&  \multirow{2}{*}{\shortstack[c]{Image\\ Alignment} }
& \multirow{2}{*}{\shortstack[c]{Text\\ Alignment }  }
& \multirow{2}{*}{\shortstack[c]{Image\\ Alignment} }
& \multirow{2}{*}{\shortstack[c]{Text\\ Alignment}  }
& \multirow{2}{*}{\shortstack[c]{Image\\ Alignment} }
& \multirow{2}{*}{\shortstack[c]{Text\\ Alignment}  }
& \multirow{2}{*}{\shortstack[c]{Image\\ Alignment} }\\ \\
\midrule

 \multirow{1}{*}{\textbf{\shortstack[c]{Single Subject}}} 
&  \multirow{1}{*}{\shortstack[c]{ 18.83$\%$}   }  & \multirow{1}{*}{\shortstack[c]{ 26.17$\%$}  }  &
\multirow{1}{*}{\shortstack[c]{ 26.17$\%$}  } &
\multirow{1}{*}{\shortstack[c]{ \textbf{28.00$\%$}}   } & 
\multirow{1}{*}{\shortstack[c]{ 26.83$\%$}} &
\multirow{1}{*}{\shortstack[c]{ 20.67$\%$}   } &
\multirow{1}{*}{\shortstack[c]{ \textbf{28.17$\%$}}   } &
\multirow{1}{*}{\shortstack[c]{ 25.17$\%$}   }\\ 

\midrule

\multirow{1}{*}{\shortstack[c]{\textbf{Two Subjects} } } &

\multirow{1}{*}{\shortstack[c]{ 14.83$\%$   }}  & \multirow{1}{*}{\shortstack[c]{ 18.17$\%$ }}  &
\multirow{1}{*}{\shortstack[c]{ 25.33$\%$  }} &
\multirow{1}{*}{\shortstack[c]{ 25.67$\%$   }} & 
\multirow{1}{*}{\shortstack[c]{ 29.00$\%$  }} &
\multirow{1}{*}{\shortstack[c]{ 27.00$\%$   }} &
\multirow{1}{*}{\shortstack[c]{ \textbf{30.83$\%$}  }} &
\multirow{1}{*}{\shortstack[c]{ \textbf{29.17$\%$}  }}\\
\midrule

\multirow{1}{*}{\shortstack[c]{\textbf{Three Subjects} } } &

\multirow{1}{*}{\shortstack[c]{ 10.67$\%$   }}  & \multirow{1}{*}{\shortstack[c]{ 12.00$\%$  }}  &
\multirow{1}{*}{\shortstack[c]{ 24.67$\%$  }} &
\multirow{1}{*}{\shortstack[c]{ 23.50$\%$   }} & 
\multirow{1}{*}{\shortstack[c]{ 30.83$\%$   }} &
\multirow{1}{*}{\shortstack[c]{ 30.16$\%$  }} &
\multirow{1}{*}{\shortstack[c]{ \textbf{34.17$\%$}   }} &
\multirow{1}{*}{\shortstack[c]{ \textbf{34.33$\%$}   }}\\
\midrule

\multirow{1}{*}{\shortstack[c]{\textbf{Four Subjects} } } &

\multirow{1}{*}{\shortstack[c]{ 8.83$\%$   }}  & \multirow{1}{*}{\shortstack[c]{ 7.83$\%$  }}  &
\multirow{1}{*}{\shortstack[c]{ 20.17$\%$  }} &
\multirow{1}{*}{\shortstack[c]{ 22.67$\%$   }} & 
\multirow{1}{*}{\shortstack[c]{ 33.17$\%$   }} &
\multirow{1}{*}{\shortstack[c]{ 34.17$\%$  }} &
\multirow{1}{*}{\shortstack[c]{ \textbf{37.83$\%$}   }} &
\multirow{1}{*}{\shortstack[c]{ \textbf{35.33$\%$}  }}\\
\bottomrule
\vspace{-10pt}
\end{tabular}
}
\vspace{-0.35cm}
\caption{User study results. The value represents the percentage of users that think the image generated by the corresponding method is the best. The results show that our method is the most preferred by users for multi-subject generation, on both image and text alignment.
}\vspace{-0.1cm}

\label{tbl:human_eval}
\vspace{-18pt}
\end{table*}

%% file: section/7.3.more_results.tex
\section{More Results}\label{sec:more_results}
\subsection{Sequential Training Comparison.}
\begin{figure}[t]
  \centering
    \begin{overpic}[width=0.99\linewidth]{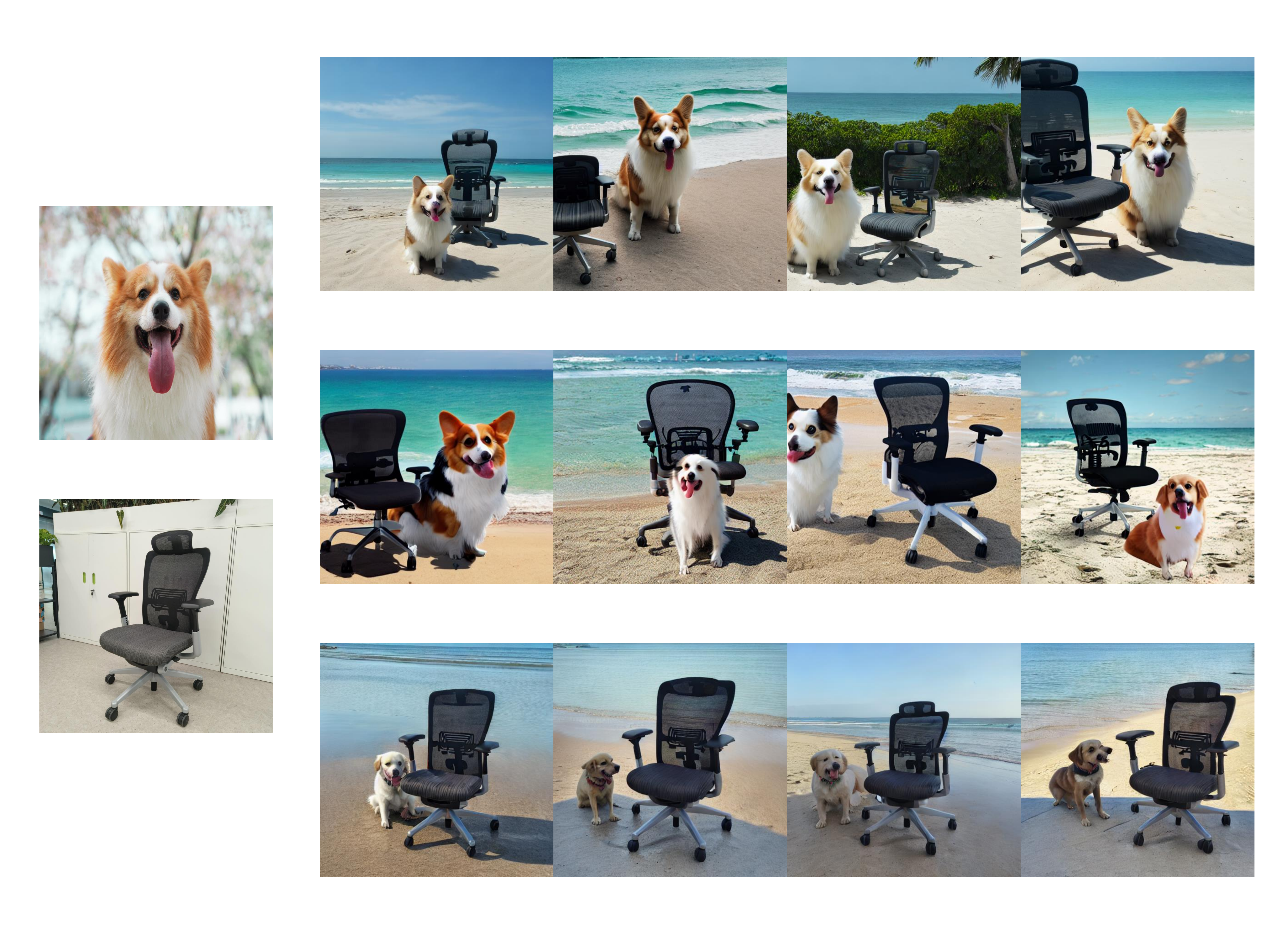}
  \put(7.5,-1){\small{Target Images}}
  \put(56,70){\small{Cones (Ours)}}
  \put(56,47){\small{Custom Diffusion}}
  \put(57,24.5){\small{Dreambooth}}
  \put(56,-1){\small{Generated images}}
  \end{overpic}
  \caption{Sequential training results. The model learns "dog" and "chair" sequentially. It can be seen that the other two methods have severe forgetting of "dog". Cones retains better for both subjects.}
  \vspace{-0.1in}
  \label{fig:seq}
  \vspace{-0.1in}
\end{figure}

 As shown in~\ref{fig:seq}, we also evaluate Cones of sequential training on two subjects. Specifically, we optimize the concept-implanting loss for the second subject while shutting the corresponding concept neurons of the first subject. In the case of sequential training, we observe severe forgetting of the first concept for Custom Diffusion and DreamBooth, while Cones performs much better.

\subsection{Style Conversion.}
\begin{figure*}[t]
  \centering
    \begin{overpic}[width=0.96\linewidth]{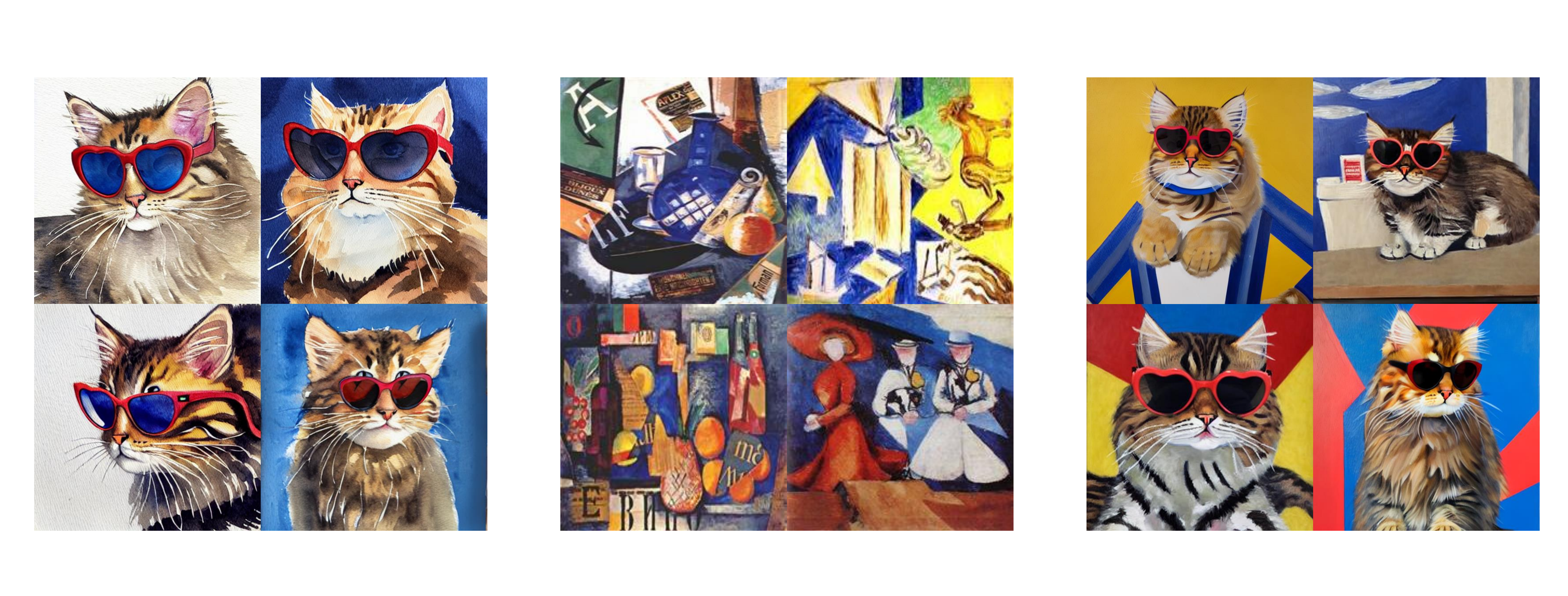}
  \put(9,1){\small{Generated subject}}
  \put(45,1){\small{Style subject}}
  \put(76,1){\small{Impainting results}}
    \put(2.5,35){\small{\makecell{A watercolor painting of a V$_{2}$* cat \\
    wearing a V$_{3}$* sunglasses.}}}
    \put(47,35){\small{\makecell{V$_{1}$* art}}}
    \put(71,35){\small{\makecell{A V$_{1}$* art painting of a V$_{2}$* cat \\ wearing a V$_{3}$* sunglasses.}}}
  \end{overpic}
  \caption{Style conversion results. The first column is the style conversion of the image through the knowledge of the pretrained model, the second column is a specific style, and the third column is the result of our generation in the specific style.}
  \label{fig:supp_impaint}
  \vspace{0.1in}
\end{figure*}

As shown in~\cref{fig:supp_impaint}, Cones is able to express a certain style through text guidance and can also be fine-tuned on a fixed style.

\subsection{Editing Performance}
\begin{figure*}[t]
  \centering
    \begin{overpic}[width=0.96\linewidth]{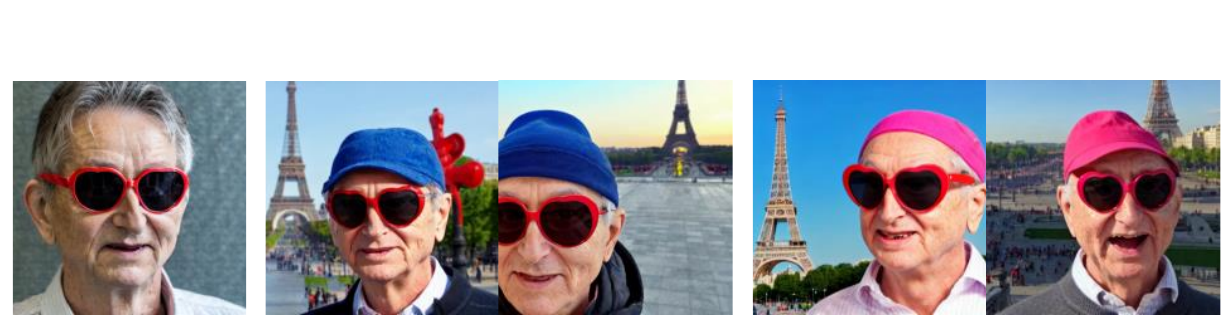}
  \put(2,23){\small{\makecell{A V$_{1} $* man with \\ V$_{2}$* sunglasses on face.}}}
  \put(3,-1){\small{Generated subject}}
  \put(54,-1){\small{Edited subject}}
    \put(22,23){\small{\makecell{A V$_{1}$* man is smiling, with V$_{2}$* sunglasses \\ on face,  with blue hat on head, \\ with Eiffel Tower in the background.}}}
    \put(62,23){\small{\makecell{A V$_{1}$* man is laughing, with V$_{2}$* sunglasses \\ on face,  with pink hat on head, \\ with Eiffel Tower in the background.}}}
  \end{overpic}
  \caption{Editing results.}
  \label{fig:supp_hinton}
  \vspace{-0.1in}
\end{figure*}

As shown in~\ref{fig:supp_hinton}, Cones can capture more similarity in the generation results with the textual descriptions when editing images with text prompts, like  expression switching and changing object colors, as well as adding a background. 
\subsection{Overfitting on the training prompt template.}
During fine-tuning, the target images are trained with the text prompt "photo of a V$_{1}$* {class}", where V$_{1}$* is the text identifier of the subject, and {class} is the class to which the subject belongs. As mentioned in Custom Diffusion~\cite{cosdiff}, after fine-tuning the models, the generations shift towards the target images and have less diversity compared to the pretrained model with the prompt "photo of a V$_{1}$* {class}". However, as shown in figure~\cref{fig:overfitting}, the images generated by Cones have more diversity, which proves that Cones alleviates overfitting.
\subsection{More results on multi subjects}
As mentioned in Custom Diffusion~\cite{cosdiff}, the pretrained model encounters difficulty generating multiple subjects described in a single text prompt. As shown in~\cref{fig:more_multi}, when Cones incorporates the Attend-and-Excite~\cite{feng2022training,chefer2023attend} method to address this issue, it generates better results.